\documentclass[10pt,twocolumn,letterpaper]{article}

\usepackage{Style_Ref/wacv}
\usepackage{times}
\usepackage{epsfig}
\usepackage{graphicx}
\usepackage{amsmath}
\usepackage{amssymb}

\usepackage{float}
\usepackage[linesnumbered,ruled]{algorithm2e}
\usepackage{comment}
\usepackage{xcolor}
\usepackage{enumitem}
\usepackage{courier}

\usepackage{makecell,interfaces-makecell}
\usepackage{arydshln}
\usepackage{subcaption}
\usepackage{wrapfig}
\usepackage{amsfonts} 
\DeclareMathSymbol{\shortminus}{\mathbin}{AMSa}{"39}

\usepackage{multirow}

\usepackage[pagebackref=true,breaklinks=true,colorlinks,bookmarks=false]{hyperref}

\def\wacvPaperID{506} 

\wacvfinalcopy 

\ifwacvfinal
\def\assignedStartPage{1} 
\fi

\ifwacvfinal
\usepackage[breaklinks=true,bookmarks=false]{hyperref}
\else
\usepackage[pagebackref=true,breaklinks=true,colorlinks,bookmarks=false]{hyperref}
\fi

\ifwacvfinal
\else
\fi

\title{Class-Discriminative CNN Compression}

\author{%
  Yuchen~Liu, David~Wentzlaff, S.Y.~Kung
    \\
  Princeton University\\
  \texttt{\{yl16, wentzlaf, kung\}@princeton.edu} \\
}

\begin{document}
\maketitle
\newcommand\metric{G-SD}
\newcommand\method{CDC}
\newcommand\analysis{FLOP-normalized sensitivity analysis}
\newcommand\feamap{\mathcal{F}}

\begin{abstract}
Compressing convolutional neural networks (CNNs) by pruning and distillation has received ever-increasing focus in the community.
In particular, designing a class-discrimination based approach would be desired as it fits seamlessly with the CNNs training objective.
In this paper, we propose class-discriminative compression (\method), 
which injects class discrimination in both pruning and distillation to facilitate the CNNs training goal.
We first study the effectiveness of a group of discriminant functions for channel pruning,
where we include well-known single-variate binary-class statistics like Student's T-Test in our study via an intuitive generalization.
We then propose a novel layer-adaptive hierarchical pruning approach, 
where we use a coarse class discrimination scheme for early layers and a fine one for later layers.
This method naturally accords with the fact that CNNs process coarse semantics in the early layers and extract fine concepts at the later.
Moreover, we leverage discriminant component analysis (DCA) to distill knowledge of intermediate representations in a subspace with rich discriminative information,  
which enhances hidden layers' linear separability and classification accuracy of the student. 
Combining pruning and distillation, \method\ is evaluated on CIFAR and ILSVRC-2012, where we consistently outperform the state-of-the-art results.

\end{abstract}
\section{Introduction}
\label{sec:intro}

Convolutional neural networks (CNNs) have become a mainstream machine learning model for various computer vision tasks, 
such as image classification~\cite{ simonyan2014very, he2016deep}, 
object detection~\cite{girshick2015fast,renNIPS15fasterrcnn}, and 
semantic segmentation~\cite{long2015fully, ronneberger2015u}.
To gain better recognition performance, a popular approach is to grow deeper and wider models.
However, such CNNs require a larger storage space and higher computational cost, 
making them unsuitable for edge devices like mobile phones and embedded sensors. 

Many methods have been proposed for CNN compression. 
For example: weight quantization~\cite{chen2015compressing,courbariaux2016binarized}, 
tensor low-rank factorization~\cite{jaderberg2014speeding,lebedev2014speeding},  
network pruning~\cite{han2015learning, han2015deep, zhuang2018discrimination, he2019filter}, 
and knowledge distillation~\cite{hinton2015distilling,romero2014fitnets}.  
Among them all, a combination of channel pruning and knowledge distillation is the preferable method to learn smaller dense models, 
which can easily leverage Basic Linear Algebra Subprograms (BLAS) libraries~\cite{li2016pruning}.

While CNNs are fundamentally trained to differentiate objects from different classes,
the study of discrimination based network compression is quite limited.
Prior class-discriminative pruning works~\cite{molchanov2017pruning,zhuang2018discrimination,lin2018accelerating,kung2019methodical,zhonghui2019gate} lack effectiveness study for their pruning metrics,
where they propose and evaluate their metrics singly without comparing to other well-known discriminant functions, like Maximum Mean Discrepancy~\cite{gretton2012kernel}.
Besides, these works ignore the hierarchical nature of CNNs' semantic extraction and only use fine class discrimination for both early and later layers, which could be sub-optimal.
For knowledge distillation, 
while the output distillation scheme~\cite{hinton2015distilling,yun2020regularizing} normally incorporates class discrepancy knowledge,
intermediate discriminative distillation has rarely been attempted.

\begin{figure}[t]
    \centering
    \includegraphics[width=0.48\textwidth]{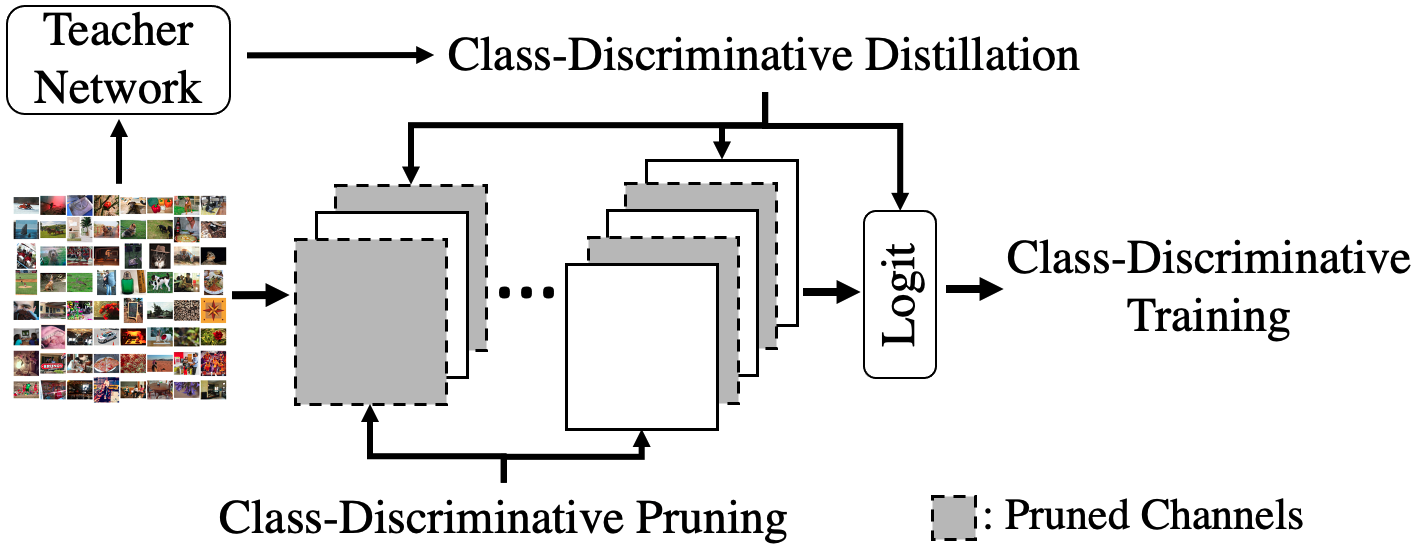}
    \vspace{-0.5cm}
    \caption{We propose a novel compression scheme, class-discriminative compression (CDC),
    which leverages class discrepancy for channel pruning and knowledge distillation, fitting seamlessly with CNN class-discriminative training. 
    }
    \label{fig:intro_idea}
    \vspace{-0.5cm}
\end{figure}


To this end, we propose a novel approach to compress classification CNNs, dubbed class-discriminative compression (CDC), in Fig.~\ref{fig:intro_idea}. 
We design a unified framework for class-discriminative training, pruning, and distillation, 
which all aim to improve the final recognition performance.

We first study a group of closed-form functions to find the best metric for class-discriminative channel pruning.
This group includes high-dimensional metrics like Maximum Mean Discrepancy (MMD)~\cite{gretton2012kernel}
and single-variate binary-class statistics like Student's T-Test~\cite{lehmann2006testing}, 
for which we provide an intuitive and lightweight generalization for the high-dimensional multi-class channel scoring. 
Surprisingly, a generalized metric, generalized Symmetric Divergence (\metric), achieves the best pruning results.
We then propose a novel hierarchical pruning paradigm, 
which uses a coarse class granularity to evaluate class discrepancy for  channels at front layers, 
and a fine granularity for rear-layer channels.
This adaptation is based on the fact that CNNs extract coarse semantics at early layers while understand fine concepts at the later~\cite{zeiler2014visualizing}.
In our study, we find that the hierarchical scheme further improves the pruning results.

Moreover, we make the first attempt to design a subspace distillation approach to allow the class-discriminative knowledge concentratedly distilled to the student model at intermediate layers.
To achieve that, we use 
discriminant component analysis (DCA)~\cite{kung2017discriminant}, 
which analytically derives linear weights that transform the layer into a subspace with the most class-discriminative power. 
This scheme improves student hidden layers' linear separability and achieves a better classification accuracy.

Our contributions are summarized as follows: 

(1) We propose a novel framework, 
class-discriminative compression (\method), to learn efficient CNNs. 
This framework incorporates class discrepancy in the process of channel pruning and knowledge distillation, 
which is naturally coherent with the discriminative training objective.  
(2) We study  a group of discriminant functions for discriminative pruning, 
and propose a layer-adaptive hierarchical pruning scheme, 
which measures the class discrepancy in different label granularities based on layers' positions.
(3) We make the first attempt to distill intermediate knowledge in the subspace with the most class-discriminative power,
which enhances linear separability of student's hidden layers and achieves better distillation performance.
(4) We demonstrate the advantage of \method\ on CIFAR and ILSVRC-2012 with VGG, ResNet, and MobileNet-V2.
On ILSVRC-2012, our compressed ResNet-50 achieves a top-1 accuracy of 76.89\% (0.04\% accuracy gain from the baseline) with 44.3\% FLOPs reduction, 
outperforming state of the arts.

\section{Related Work}

\textbf{Channel Pruning.}
Channel pruning is a promising approach to enhance  network efficiency~\cite{yu2018nisp, ye2018rethinking, he2019filter,kung2019methodical,he2020learning,guo2020dmcp,lin2020hrank,chin2020towards}.
Some works leverage norm statistics of weight parameters~\cite{li2016pruning,liu2017learning,he2019filter},
feature maps reconstruction losses~\cite{he2017channel,luo2017thinet}, 
ranks of the feature maps~\cite{lin2020hrank}, 
and self-learned importance coefficients~\cite{huang2018data,chin2020towards} to evaluate channel redundancy, 
without the use of discriminative information. 

In line with our work, 
several discrimination-based pruning methods are proposed~\cite{molchanov2017pruning,zhuang2018discrimination, lin2018accelerating, zhonghui2019gate,kung2019methodical}.
Molchanov et al.~\cite{molchanov2017pruning}, Lin et al.~\cite{lin2018accelerating}, and Zhong et al.~\cite{zhonghui2019gate} use Taylor expansion to estimate the accuracy/entropy loss caused by removing a channel.
They omit high order terms, making the evaluation less accurate,
and require back propagation of class information from the outputs to early layers,
where the inter-layer dependency could affect channel saliency measurement.
On the contrary, we adopt closed-form metrics to directly evaluate channels' activations discrepancy without approximation and back-propagation.
Zhuang et al.~\cite{zhuang2018discrimination} and Kung et al.~\cite{kung2019methodical} use a entropy loss and a closed-form discriminant function to directly evalaute channel activations discrepancy,
but require time-consuming iterative optimization and heavy matrix operations. 
In contrast, 
our metric doesn't need iterative steps and expensive matrix computation,
which is more time efficient.

Moreover, these studies solely focus on a single discriminant loss/metric,
whilst we provide an analysis on the effectiveness of a group of closed-form metrics. 
We also introduce a novel hierarchical pruning paradigm where the class discrimination is measured on different label granularities for early and later layers.
Such adaption of label granularity in channel pruning fits seamlessly with the nature of coarse to fine semantic understanding in CNNs.

\textbf{Knowledge Distillation.}
The idea of knowledge distillation
is pioneered by Hinton et al.~\cite{hinton2015distilling} to allow a student
classifier to mimic the output of its teacher by a soft label entropy loss. 
Romero et al.~\cite{romero2014fitnets} propose a popular hint-layer method for intermediate distillation (similar methodology adopted in~\cite{chen2017learning,li2020few}),
which uses gradient-learned weights to linearly transform student's hidden layers to the same depth as the teacher's and minimize their norm-based loss, 
while no classification information is distilled.
Moreover, the full space of the layer could contain redundant knowledge for classification, making~\cite{romero2014fitnets} less effective. 
In our work, we propose to distill classification information in the subsapce of hidden layers by discriminant component analysis (DCA)~\cite{kung2017discriminant}, 
where we achieve better distillation results. 
There is also also an attention transfer approach for intermediate distillation~\cite{zagoruyko2016paying}, yet again no supervised class information is distilled.
To the best of our knowledge, only Li et al.~\cite{li2020local} propose a discriminative intermediate distillation scheme by extracting local patterns in the feature maps. 
However, this method requires a significant amount of overhead to train the class-wise attention module, 
which could hinder its usage on large high-performing teacher networks, e.g., ResNet-50.
On the contrary, our subspace distillation scheme is more scalable that allows us to use the full-size ResNet-50 as the teacher.

\textbf{Class-Discriminative Analysis.}
Our work is also closely related to techniques for class-discriminative analysis,
such as high-dimensional class-discriminative functions of Discriminant Information (DI)~\cite{kung2019methodical} and Maximum Mean Discrepancy (MMD)~\cite{gretton2012kernel}. 
We also include a set of single-variate binary-class discriminant metrics, Student's T-Test (Ttest)~\cite{lehmann2006testing}, 
Absolute SNR (AbsSNR)~\cite{golub1999molecular},
Symmetric Divergence (SD)~\cite{mak2006solution}, 
and Fisher Discriminant Ratio (FDR)~\cite{pavlidis2001gene} in our study.
These metrics are originally defined to measure the significance of two class's difference for a univariate dataset and have been adopted for machine learning tasks. 
For example, SD is used to select discriminative individual features in bioinformatics feature vectors for dimension reduction and efficient classification~\cite{mak2006solution}. 
However,
no prior work has applied them under the context of channel pruning, 
and we take the first step to generalize them for high-dimensional multi-class channel scoring.

Discriminant component analysis (DCA)~\cite{kung2017discriminant} also plays a key role in our work.
It can be seen as a multi-class linear discriminant analysis (LDA) 
and a supervised version of principle component analysis (PCA).
It finds components of a data matrix that can transform it into a subspace with the most class linear separability. 
We apply DCA to achieve class-discriminative distillation at intermediate layers.

\section{Methodology}
 
\subsection{Discriminant Functions}\label{sec:methodology_metrics}
 
\begin{figure}[t]
 	\centering
  \includegraphics[width=0.47\textwidth]
  {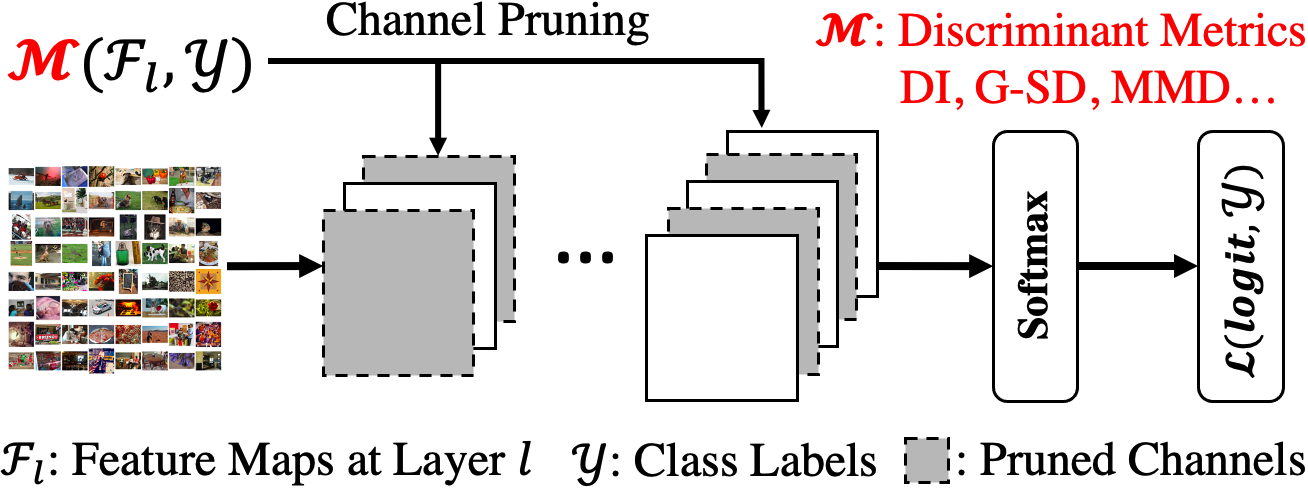}
    \vspace{-0.3cm}
        \caption{Finding the best class discriminant metrics for channel pruning.}\label{fig:discrimint_function_cp}
   \vspace{-0.5cm}
 \end{figure}

To evaluate the class discrepancy of a channel, we formulate it as $\mathcal{M}(\feamap, \mathcal{Y})$, 
where $\mathcal{M}$ is the discriminant metric, $\feamap$ is the set of feature maps obtained at a channel, 
and $\mathcal{Y}$ is the label numbering scheme. 
Although discriminant metrics are all mathematically well-defined, their empirical effectiveness on channel pruning remains understudied. 
As shown in Fig.~\ref{fig:discrimint_function_cp}, the very first thing we want to know is which discriminant metric works best for channel pruning.

We study a group of closed-form metrics including MMD and DI, which can be applied on feature maps directly without back-propagation and iterative optimization.
We also generalize four univariate binary-class metrics, T-test, AbsSNR, FDR, and SD, for channel pruning.
We use SD~\cite{kung2014kernel} as an example to illustrate our generalization method.
Let us denote an $n$-sample $2$-class single-variate dataset as $\mathcal{D} = \{ (x_i,b_i) \}_{i=1}^n$, 
where $b_i \in \{0,1\}$ is the binary labeling scheme $\mathcal{B}$. 
Let $\mathcal{D}^+ = \{ x_i~|~(x_i,y_i)\in \mathcal{D},~y_i=1 \}$ and 
$\mathcal{D}^- = \{ x_i~|~(x_i,y_i)\in \mathcal{D},~y_i=0 \}$ denote two partitions of $\mathcal{D}$ based on $\mathcal{B}$, 
SD of $\mathcal{D}$ is defined as:
\vspace{-0.06cm}
\begin{equation}\label{eqn:SD}
   \mathrm{SD}(\mathcal{D}, \mathcal{B}) = \frac{1}{2} \left( 
\frac{\sigma_{\mathcal{D}^+}^2}{\sigma_{\mathcal{D}^-}^2} + \frac{\sigma_{\mathcal{D}^-}^2}{\sigma_{\mathcal{D}^+}^2}
\right) 
+ 
\frac{1}{2}
\left( 
\frac{ (\mu_{\mathcal{D}^+} - \mu_{\mathcal{D}^-})^2 }
{\sigma_{\mathcal{D}^+}^2 + \sigma_{\mathcal{D}^-}^2}
 \right) - 1 
\end{equation}
where $\mu_{\mathcal{D}^+}$ and $\sigma_{\mathcal{D}^+}^2$ are the sample mean and variance of 
${\mathcal{D}^+}$, and $\mu_{\mathcal{D}^-}$ and $\sigma_{\mathcal{D}^-}^2$ are the statistics for ${\mathcal{D}^-}$.

For an $N$-sample $Y$-class dataset, 
we denote the feature maps of a CNN channel as $\mathcal{F} = \{(f_i, y_i)\}^N_{i = 1}$, 
where $f_{i} \in \mathbb{R}^{W \times H}$ denotes the feature map of the $i$-th input image, 
$y_i \in [1\textit{:}Y]$ is the $i$-th class label,
and $W/H$ are the spatial sizes. 
We first partition $\feamap$ as $\feamap^c$ and  $\feamap^{-c}$ where 
$\feamap^c = \{f_{i}~|~(f_{i}, y_i) \in \feamap,~y_i = c \}$ and 
$\feamap^{-c} = \{f_{i}~|~(f_{i}, y_i) \in \feamap,~y_i \neq c \}$, $\forall~c \in [1\textit{:}Y] $.  
By this partition, denoted as $\mathcal{B}_c$, we can find the statistics in Eqn.~\ref{eqn:SD} in a two-class manner. 
Note each $f_{i}$ in $\feamap^c$ is a 2D feature map with $W \times H$ activations, 
and thus there are $|\feamap^c| \times W \times H$ activations in $\feamap^c$ in total. 
We then define two statistics operators $g_{mean}$ and $g_{var}$ on $\feamap^c$, which return the mean and variance over these $|\feamap^c| \times W \times H$ activations.
We thus get $\mu_c = g_{mean}(\feamap^c )$ and $\sigma_c^2= g_{var}(\feamap^c )$ for $\feamap^c$, and their
counterparts $\mu_{-c}$ and $\sigma_{-c}^2$.
The SD score for $\mathcal{B}_c$ is thus:
\vspace{-0.15cm}
\begin{equation}\label{eqn:l_class_sd}
   \mathrm{SD}(\feamap, \mathcal{B}_c) = \frac{1}{2} \left( 
\frac{\sigma_c^2}{\sigma_{-c}^2} + \frac{\sigma_{-c}^2}{\sigma_c^2}
\right) 
+ 
\frac{1}{2}
\left( 
\frac{ (\mu_c - \mu_{-c})^2 }
{\sigma_c^2 + \sigma_{-c}^2}
 \right) - 1 
\end{equation}
$\mathrm{SD}(\feamap , \mathcal{B}_c)$ captures the discriminativenesss of class $c$ relative to the other classes in $\feamap$.
In general, we want to select channels that distinguish all classes well on average.
Thus, under the $Y$-class setting, the generalized Symmetric Divergence (\metric) of $\feamap$ is formally defined as:
\vspace{-0.3cm}
\begin{equation}\label{eqn:SymDiv}
   \mathrm{G{\shortminus}SD}(\feamap, \mathcal{Y}) = \frac{1}{Y}\sum_{c = 1}^Y {\mathrm{SD}(\feamap , \mathcal{B}_c)}
   \vspace{-0.3cm}
\end{equation}
Such generalization method is applicable to other single-variate binary-class metrics, 
and incurs no expensive operations (e.g., matrix inversion, SVD),
which makes the generalized metrics scalable to large networks and datasets.

\subsection{Hierarchical Pruning}\label{sec:hierachical_pruning}

\begin{figure}[t]
 	\centering
  \includegraphics[width=0.47\textwidth]{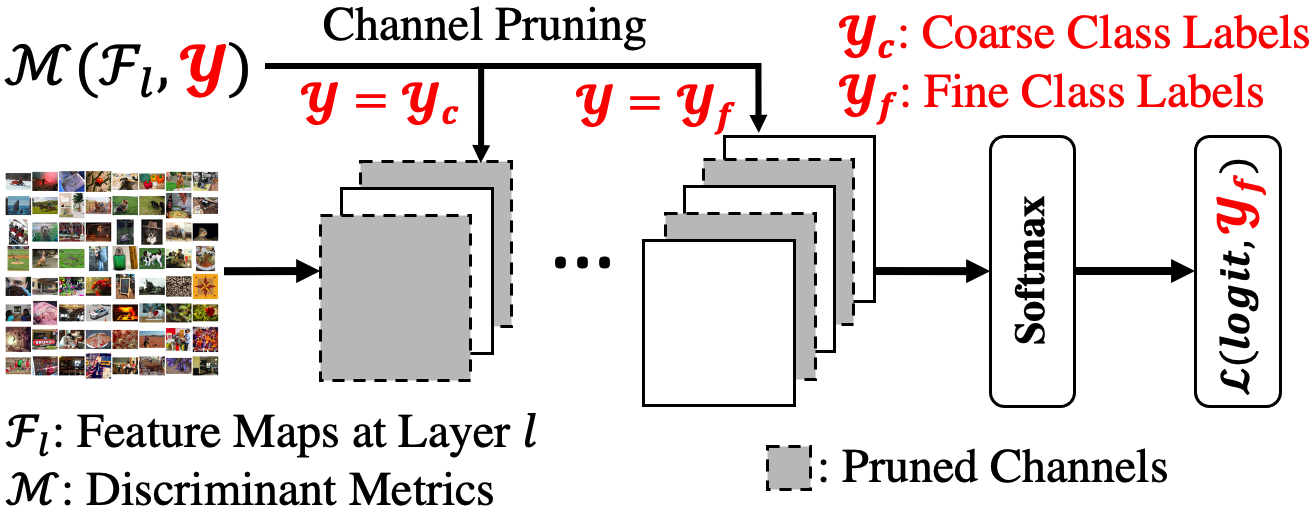}
  \vspace{-0.3cm}
        \caption{Adapting the label granularity for channels' class discrepancy measurement based on layer positions.}\label{fig:hierarchical_pruning}
        \vspace{-0.4cm}
 \end{figure}

In our formulation $\mathcal{M}(\feamap, \mathcal{Y})$,
$\feamap$ is determined by input images and network's weights, 
while $\mathcal{Y}$ can be calibrated for different layers.
After finding out the best metric $\mathcal{M}$, we investigate the settings of $\mathcal{Y}$ for more effective discriminative pruning.
It is widely recognized that CNN learns coarse semantics (fruit, vehicle) in the early layers
while extracting finer class (apple, truck) concepts in the later layers~\cite{zeiler2014visualizing}.
Inspired by the nature of hierarchical semantic separation in CNN, 
we propose to adapt the granularity of $\mathcal{Y}$ based on layer positions for class discrepancy measurement  as shown in Fig.~\ref{fig:hierarchical_pruning}.
Specifically, we define a watershed layer $l_{WS}$, 
where we evaluate the channel discrepancy by a coarse class labeling scheme $\mathcal{M}(\feamap, \mathcal{Y}_c)$ when $l \leq l_{WS}$, 
and use fine label pruning $\mathcal{M}(\feamap, \mathcal{Y}_f)$ when $l > l_{WS}$.

While most image datasets only provide fine class labels $\mathcal{Y}_f \in [1\text{:}F]$, 
the coarser labeling scheme $\mathcal{Y}_c \in [1\text{:}C]$ need to be derived on our own.
To tackle this issue, we use a pretrained CNN $Net$  to group similar fine categories into the same coarse category, i.e.,
to learn a disjoint many-to-one mapping $Q: [1\text{:}F] \rightarrow [1\text{:}C]$.
We investigate two methods.

\textbf{Clustering on Class Centroids.} 
We randomly sample a held-out set of images from the training set. 
We feed these images through $Net$ and get the last hidden activations. 
We calculate the activations' class centroid for each fine label, 
denoted as $\mathcal{H} = \{h_1, h_2, ..., h_F \}$,
and run a K-means clustering on $\mathcal{H}$ with $C$ clusters to get the mapping $Q$.

\textbf{Clustering on Confusion Matrix.}
We feed the held-out images through $Net$ to get their predicted labels.
Based on the predicted and true labels, we construct a confusion matrix $M \in \mathbb{R}^{F \times F}$, 
where $M_{i,j}$ denotes the number of images with true label $i$ but predicted as label $j$.
We then run a spectral clustering on $M$ with $C$ clusters which gives us $Q$.

\subsection{Intermediate Class Discrepancy Distillation}

\begin{figure}
\centering
\includegraphics[width=0.45\textwidth]{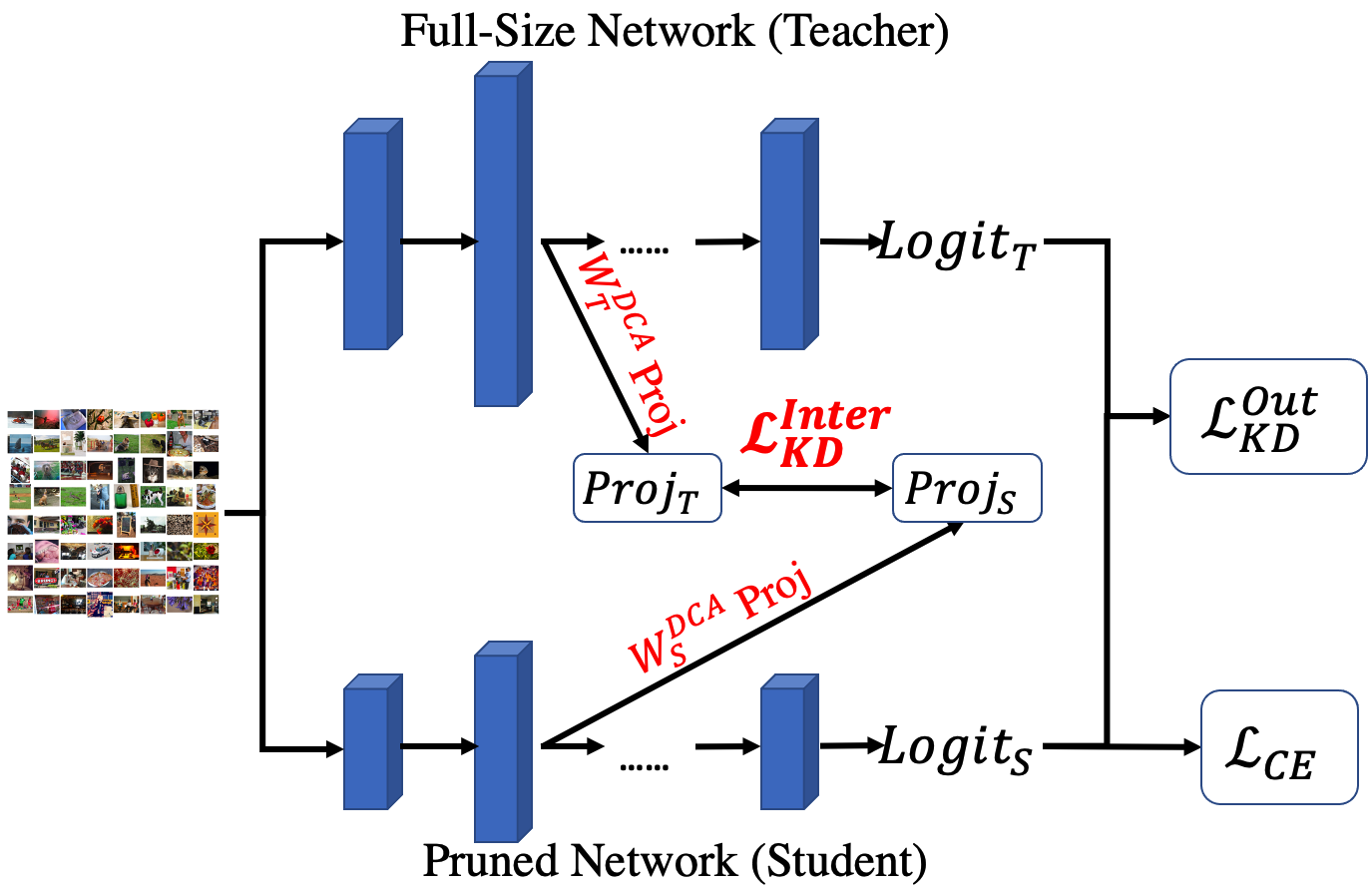}
\vspace{-0.3cm}
\caption{DCA-based intermediate distillation,
which allows class discrepancy to distill at hidden layers.}\label{fig:inter_dca_distillation}
\vspace{-0.3cm}
\end{figure}

After we obtain the pruned networks, 
we retrain them with a combined loss of cross entropy $\mathcal{L}_{CE}$ and knowledge distillation to recover their accuracies.
In particular, we propose to distill classification information at intermediate layers' subspaces found
by discriminant component analysis (DCA)~\cite{kung2017discriminant} in Fig.~\ref{fig:inter_dca_distillation}. 
We experimentally compare the DCA-based distillation with the hint layer~\cite{romero2014fitnets} in Sec.~\ref{sec:exp_dca_distill},
and find that distillation at the subspace with rich class information results in better performance.

\textbf{DCA.} Let $\mathcal{D}~\text{=}~\{(x_i, y_i)\}_{i=1}^{N}$ denote an N-sample Y-class dataset, the within-/between-class scatter matrix $\mathbf{S}_W$/$\mathbf{S}_B$:
\vspace{-0.3cm}
\begin{equation}
\mathbf{S}_W = \sum_{y = 1}^{Y}\sum_{j = 1}^{N_y}{(x_j^{(y)} - \bar{x}_y)(x_j^{(y)} - \bar{x}_y)^T}
\end{equation}
\vspace{-0.3cm}
\begin{equation}
\mathbf{S}_B = \sum_{y = 1}^{Y}{N_y(\bar{x}_y - \bar{x})(\bar{x}_y - \bar{x})^T}
\vspace{-0.3cm}
\end{equation}
where $N_y$, $x_j^{(y)}$ are the number of samples and the $j$-th sample of class $y$. $\bar{x}_y$ and $\bar{x}$ are the $y$-class centroid and overall data centroid. 
Note the center adjusted scatter matrix $\bar{\mathbf{S}} = \mathbf{S}_W + \mathbf{S}_B$. 
The discriminant components of $\mathcal{D}$ are:
\vspace{-0.3cm}
\begin{equation}~\label{eqn:DCA}
W^{DCA} = \underset{W: W^T\mathbf{S}_{W}W = I}{\arg\max}{tr(W^T\bar{\mathbf{S}}W)}
\vspace{-0.3cm}
\end{equation} 
As LDA only finds one component that best separate two classes, DCA can be seen as a multi-class version of it where $\mathbf{S}_W$ and $\mathbf{S}_B$ incoprates information for all classes.
Moreover, DCA's objective is the same as PCA while it includes an extra within-class matrix $\mathbf{S}_W$ in its constraint, 
making DCA a supervised version of PCA.
As evidenced in the original paper~\cite{kung2017discriminant}, the DCA subspace is more linearly separable than the one found by PCA.

To learn DCA weights for intermediate layers, 
we feed the held-out set of images through the network and get its intermediate activation $A \in \mathbb{R}^{N \times C \times H \times W}$, which is reshaped as $A \in \mathbb{R}^{N \times D}$, $D = C \times H \times W$. 
With a labeling scheme $\mathcal{Y}$, we apply Eqn.~\ref{eqn:DCA} to get its top $Y$ components, $W^{DCA} \in \mathbb{R}^{D \times Y}$, 
as $\mathbf{S}_B$ has a rank of $Y$.

\textbf{Distillation.}
We learn DCA for the teacher, $W^{DCA}_T$, at the start of training, 
and learn the student's DCA, $W^{DCA}_S$, every $d$ epochs.
These weights are not updated in the back-propagation, and the loss is constructed as:
\vspace{-0.3cm}
\begin{equation}
\mathcal{L}^{Inter}_{KD} = \sum_{l = 1}^{L}{||A_{T}(l)W^{DCA}_{T}(l) - A_{S}(l)W^{DCA}_{S}(l)||_1}
\end{equation}
where $A_T(l),W^{DCA}_{T}(l)$ denote the activation and DCA weights for layer $l$ of the teacher, and $A_S(l),W^{DCA}_{S}(l)$ are those for the student. 
Mathematically speaking, $\mathcal{L}^{Inter}_{KD}$ imposes regularization on student's hidden layers to push the transformed subspace as linearly separable as its teacher.

We adopt the output distillation loss $\mathcal{L}_{KD}^{Out}$ in~\cite{hinton2015distilling},
and our training loss for the student is formally defined as:
\vspace{-0.2cm}
\begin{equation}\label{eqn:training_loss}
\mathcal{L} = \mathcal{L}_{CE} + \lambda\mathcal{L}_{KD}^{Inter} + \gamma\mathcal{L}_{KD}^{Out}
\vspace{-0.2cm}
\end{equation}
where $\lambda, \gamma$ are the weights for distillation losses.

\section{Experimental Results}\label{experiment}

\subsection{Function Effectiveness Study}\label{sec:metric_effectiveness}

\begin{figure*}[t]
    \centering
    \begin{subfigure}[t]{0.48\textwidth}
        \centering
        \includegraphics[width=\textwidth]{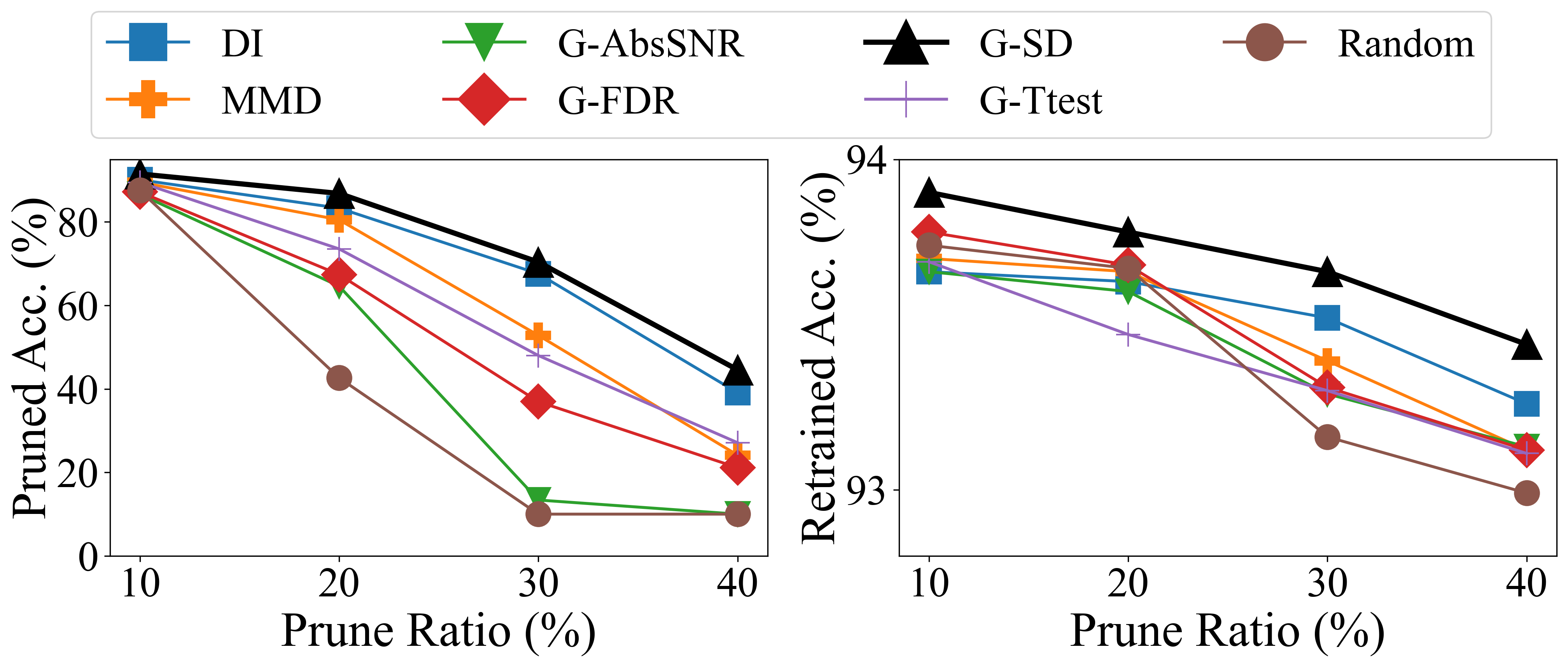}
        \vspace{-0.6cm}
        \caption{VGG-16 on CIFAR-10}
    \end{subfigure}%
    \hfill
    \begin{subfigure}[t]{0.48\textwidth}
        \centering
        \includegraphics[width=\textwidth]{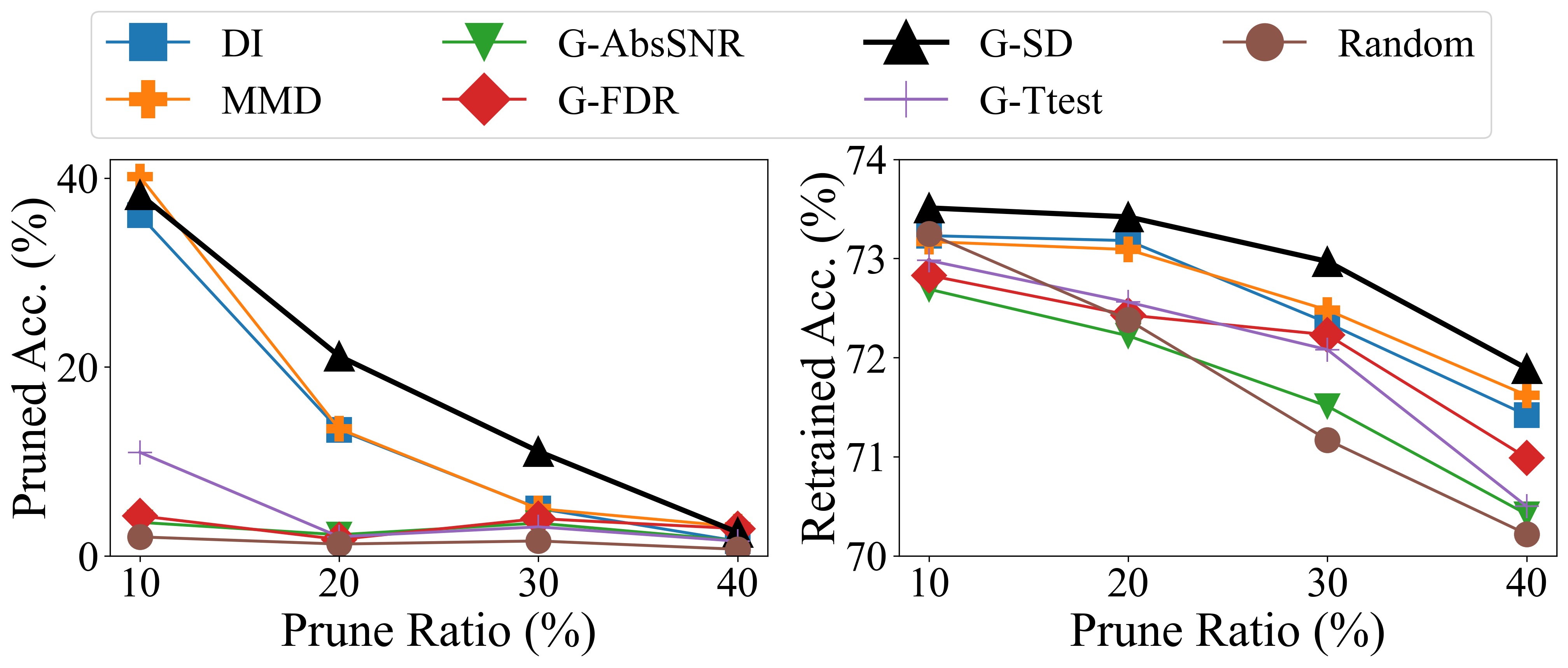}
        \vspace{-0.6cm}
        \caption{ResNet-38 on CIFAR-100}
    \end{subfigure}
    \vspace{-0.3cm}
    \caption{Empirical study on the discriminant functions' effectiveness. 
    The pool of discriminant functions are applied to prune VGG-16 and ResNet-38,
    whose pruned accuracies (without retraining) and the retrained accuracies (with retraining) are compared under different pruning ratios.
    We find a clear winner, \metric, which outperforms other metrics in all tasks.  
    }
    \label{fig:function effectiveness study}
     \vspace{-0.35cm}
\end{figure*}

We conduct one-shot pruning tests with the pool of metrics in Sec.~\ref{sec:methodology_metrics}
for VGG-16 on CIFAR-10 and ResNet-38 on CIFAR-100,
with results shown in Fig.~\ref{fig:function effectiveness study}.
For each metric, we uniformly prune 10\%, 20\%, 30\%, and 40\% of the least discriminative channels (measured by fine labels) in each layer. 
These pruned models are fine-tuned by $\mathcal{L}_{CE}$ only with the same training parameters. 
We include random pruning as the baseline. 

The discriminant metrics outperform random pruning in nearly all ratios, with and without fine-tuning, 
clearly indicating their pruning effectiveness.
We observe a leading metric, \metric, which consistently achieves the best results.
Without retraining, \metric\ outperforms DI by 5.5\% accuracy on CIFAR-10 with 40\% of channels removed
and has an 8\% accuracy gain over MMD on CIFAR-100 with 30\% of channels removed.
With retraining, \metric\ gains around 0.5\% accuracy over MMD on CIFAR-100 when 30\% of channels are removed.
Based on such consistent winning results, we adopt \metric\ in our \method\ pipeline for channel pruning.  
We also visually demonstrate the advantage of \metric\ over other state-of-the-art pruning criteria in Fig.~\ref{fig:Qualitative Channel Selection}.

\subsection{Hierarchical Pruning Study}\label{sec:hier_pruning_exp}

\begin{figure*}[t]
    \centering
    \begin{subfigure}[t]{0.3\textwidth}
        \centering
        \includegraphics[width=\textwidth]{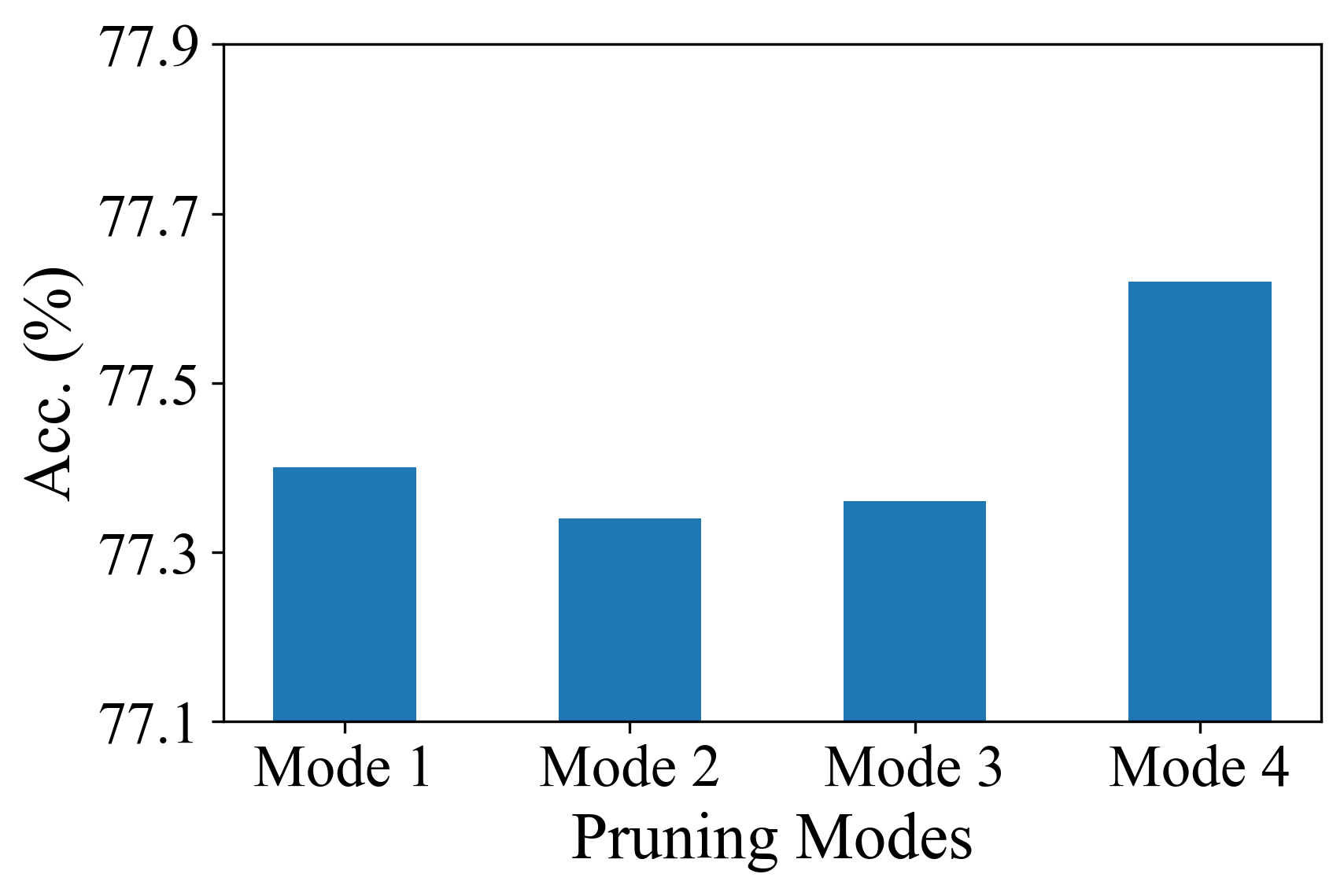}
        \vspace{-0.6cm}
        \caption{Investigation of Label Pruning Modes}\label{fig:hier_pruning_mode}
        
    \end{subfigure}
    \hfill
    \begin{subfigure}[t]{0.3\textwidth}
        \centering
        \includegraphics[width=\textwidth]{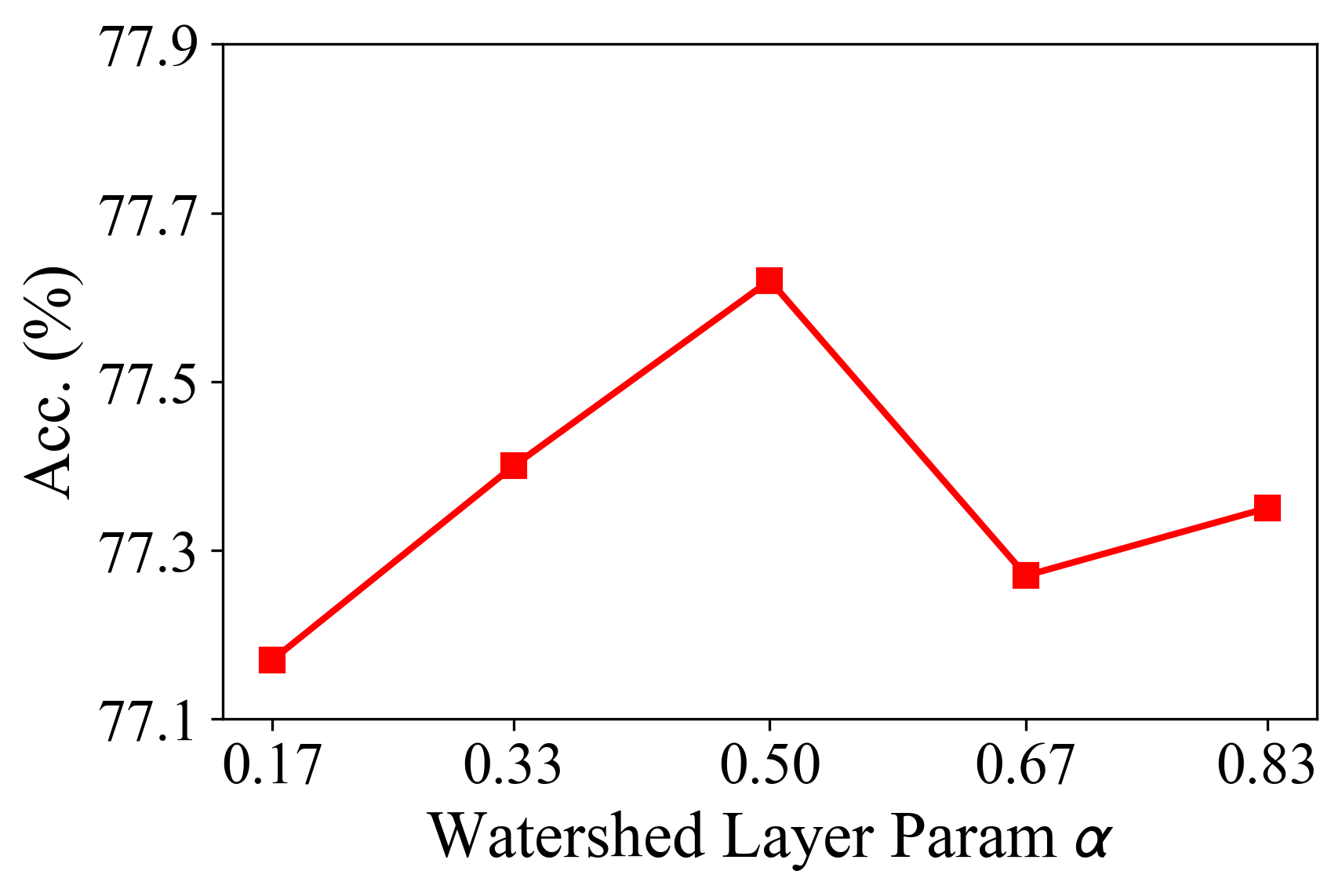}
        \vspace{-0.6cm}
        \caption{Choice of Watershed Layer}\label{fig:watershed_lay}
        
    \end{subfigure}
    \hfill
    \begin{subfigure}[t]{0.3\textwidth}
        \centering
        \includegraphics[width=\textwidth]{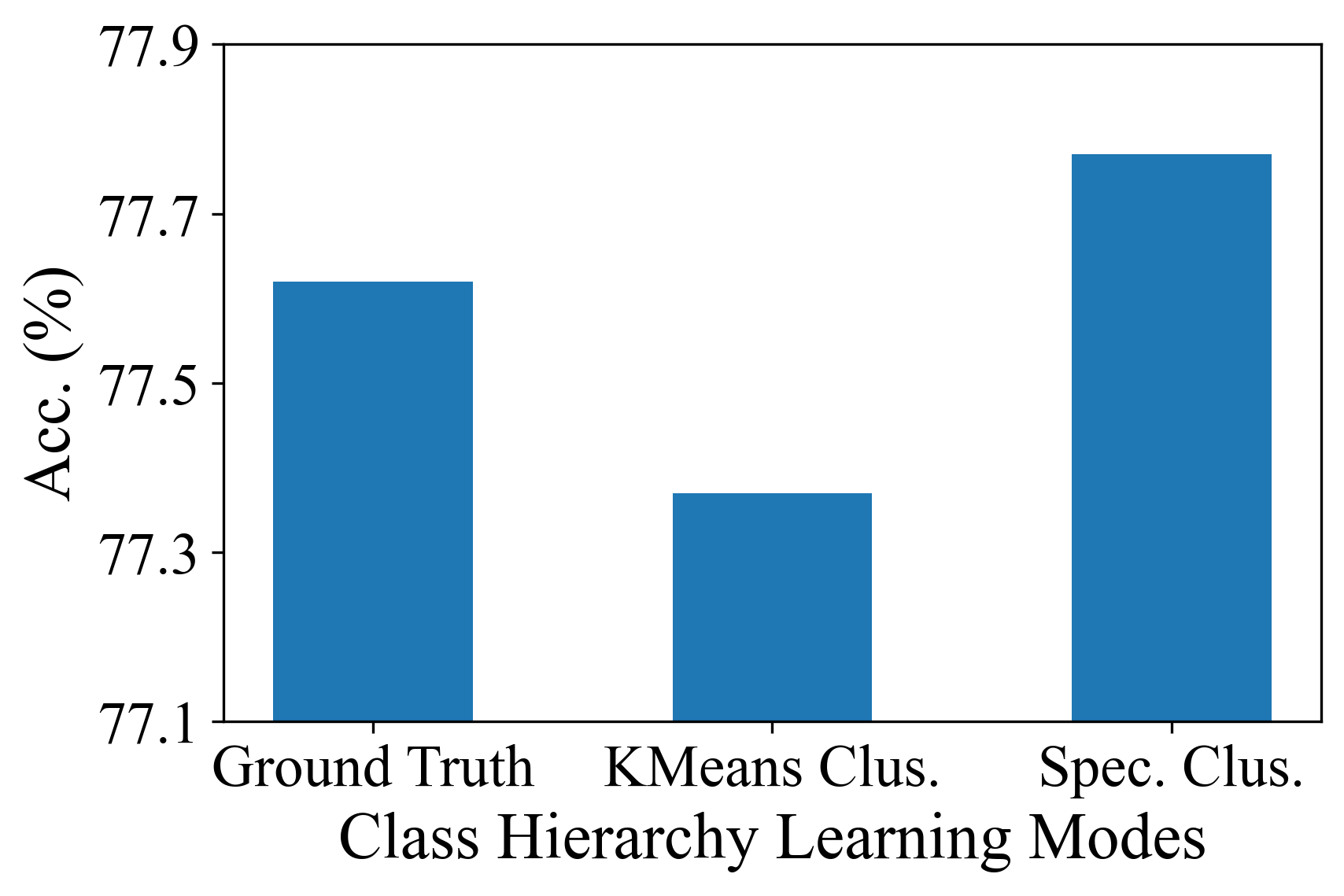}
        \vspace{-0.6cm}
        \caption{Effects of Different Class Hierarchies}\label{fig:class_hierarchies}
        
    \end{subfigure}
    \vspace{-0.3cm}
    
    \caption{Exploring different settings for hierarchical pruning (HP) with ResNet-164 on CIFAR-100. 
    (a) The front-layer $\mathcal{Y}_c$ + rear-layer $\mathcal{Y}_f$ HP scheme outperforms other label placement schemes for class-discriminative pruning.
    Notably, HP largely improves over the all-layer fine label pruning, which is used in other literatures.
    (b)  Setting $\alpha = 0.5$,
    i.e., setting the center layer as the watershed layer gives the best performance for HP.
    (c) The coarse label learned by spectral clustering on the confusion matrix outperforms the ground truth, 
    suggesting HP can be effective even without ground truth coarse labels.
    }
    \label{fig:hierarchical pruning}
    \vspace{-0.35cm}
\end{figure*}

To study the hierarchical pruning (HP) scheme, 
we use \metric\ to uniformly remove 45\% of the channels in each layer from a ResNet-164 on CIFAR-100 and retrain it by $\mathcal{L}_{CE}$ only with results in Fig.~\ref{fig:hierarchical pruning}.
CIFAR-100's ground truth coarse label can well serve as reference for the study.

\textbf{Effectiveness.} We first show the effectiveness of HP by comparing it with three other labeling schemes for class discrepancy measurement. 
All four pruning modes are:
(1) All layer fine label $\mathcal{Y}_f$ pruning.
(2) All layer coarse label $\mathcal{Y}_c$ pruning.
(3) Front layer $\mathcal{Y}_f$ + rear layer $\mathcal{Y}_c$ pruning.
(4)~Front layer $\mathcal{Y}_c$ +  rear layer $\mathcal{Y}_f$ pruning (\textbf{HP}).
We use the ground truth coarse label for $\mathcal{Y}_c$,
and set the watershed layer $l_{WS} = 0.5L$, 
where $L$ is the total number of layers in the network.
As shown in Fig.~\ref{fig:hier_pruning_mode}, 
HP (\textbf{Mode 4}) achieves the best accuracy among all schemes.
This suggests that the channels' class discrimination shall be measured based on its semantic granularity for better pruning performance.

\textbf{Watershed Layer.}  
We varies the placement of the watershed layer, 
parameterized by $l_{WS} = \alpha L, \alpha \in (0,1)$.
As shown in Fig.~\ref{fig:watershed_lay},  we find that $\alpha = 0.5$ gives the best result, 
suggesting that the CNN is processing coarse class semantics in the first half of the layers, 
and extracting the finer concepts in the second half.

\textbf{Class Hierarchies.} 
While most image datasets don't have ground truth coarse labels,
we further evaluate proposed HP using the coarse labels learned by the clustering algorithms in Sec.~\ref{sec:hierachical_pruning}, and set $l_{WS} = 0.5L$.
We set the number of learned coarse classes to be 20, which is the same as the ground truth scheme.  
As shown in Fig.~\ref{fig:class_hierarchies}, the coarse class labels learned by spectral clustering on the accuracy confusion matrix could even outperform the ground truth scheme.
This indicates that HP is generally effective, even without the ground truth coarse label.
We include a study with multiple coarse levels in Supplementary Material.

\textbf{Comparison to State of the Arts.}
\begin{table}[t]
\centering
\fontsize{8}{10}\selectfont


 \begin{tabular}{|c|c|c|c|c|c|c|}
    \hline
    Network & Method & Test Acc. (\%) & Acc. $\downarrow$  & \thead{FLOPs (M) \\ Pruned (\%)}  \\
    \hline\hline
    
    
    \multirow{4}{3em}{\thead{ResNet \\ 164}} & LCCL \cite{dong2017more} & 75.67 $\rightarrow$ 75.26 & 0.41 & 197  (21.3) \\
    
    & SLIM~\cite{liu2017learning} & 76.63 $\rightarrow$ 76.09 & 0.54 & 124 (50.6)\\
    
    & DI~\cite{kung2019methodical} & 77.63 $\rightarrow$ 76.11 & 1.52 & 105 (58.0)\\

    & {\bf HP} & {\bf 78.00 $\rightarrow$ 77.77} & {\bf 0.23} & {\bf 92 (63.2)} \\
    
    \hline
    \end{tabular}
    \vspace{-0.3cm}    
    \caption{G-SD hierarchical pruning (HP) outperforms state-of-the-art pruning methods.}\label{tab:HP_Comparison}
	\vspace{-0.4cm}

\end{table}
We further compare our G-SD HP scheme with other state-of-the-art pruning methods in Tab.~\ref{tab:HP_Comparison}, 
where our method outperforms all of them.
We achieve a 2.51\% accuracy gain over LCCL~\cite{dong2017more} with 41.9\% less FLOPs.
Compared to DI~\cite{kung2019methodical},
we achieve 1.66\% higher accuracy and 5.2\% less FLOPs.
More comparison results on ILSVRC-2012 are in Supplementary.

\subsection{Intermediate Distillation Study}\label{sec:exp_dca_distill}

\begin{figure}[t]
    \centering
    \includegraphics[width=0.42\textwidth]{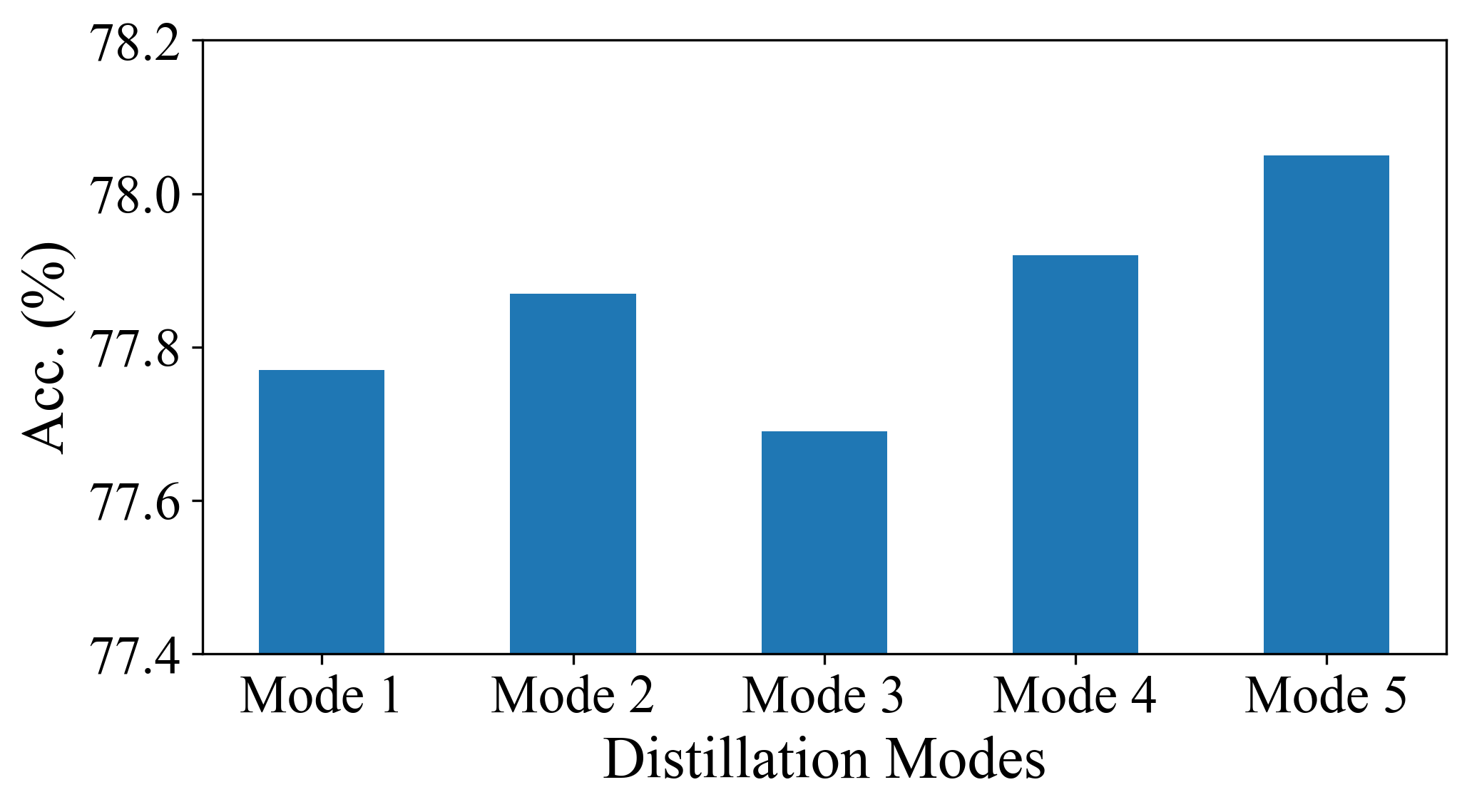}
    \vspace{-0.35cm}
    \caption{Comparing different knowledge distillation methods. 
    Adding hint-layer distillation~\cite{romero2014fitnets} (\textbf{Mode 3})  doesn't improve over output only distillation (\textbf{Mode 2}).
    On the contrary, DCA-based distillation (\textbf{Mode 4-5}) further improves the distillation performance, 
    where using the coarse label for DCA learning (\textbf{Mode 5}) achieves the best result.}
    \label{fig:kd}
        \vspace{-0.35cm}
\end{figure}

We then combine $\mathcal{L}_{CE}$ with different distillation losses to retrain HP-pruned ResNet-164 with results in Fig.~\ref{fig:kd}.
We investigate the following distillation modes which have similar computational budgets:
(1) No distillation.
(2) Only output distillation~\cite{hinton2015distilling}.
(3) Output + hint-layer intermediate distillation~\cite{romero2014fitnets}.
(4) Output + $\mathcal{Y}_f$ DCA intermediate distillation.
(5) Output + $\mathcal{Y}_c$ DCA intermediate distillation.

In this study, we set $\lambda = 10.0$ and $\gamma = 1.0$ in Eqn.~\ref{eqn:training_loss} and we only insert intermediate loss at the watershed layer for all intermediate distillation schemes. 
We include study on inserting losses at multiple intermediate layers in Supplementary Material.
We find adding hint-layer distillation (\textbf{Mode 3}) does not improve over output only distillation (\textbf{Mode 2}).
On the contrary, DCA-based distillation (\textbf{Mode 4-5}) improves the distillation quality, 
where using the coarse label $\mathcal{Y}_c$ to learn DCA weights (\textbf{Mode 5}) achieves the best results, 
which again emphasizes that we should adopt the class granularity of the layer for class discrepancy analysis. 
These results demonstrate the advantage of DCA-based intermediate distillation over the hint layer~\cite{romero2014fitnets}.

Combining HP with DCA-based distillation, 
the 2.7$\times$-accelerated ResNet-164 derived by our CDC pipeline achieves an accuracy of 78.05\% on CIFAR-100, 
further advancing the state-of-the-art compression results.

\subsection{Comparing to State of the Arts}

{\bf Benchmarks.} Combing HP and DCA-based distillation, we evaluate \method\ with VGGNet~\cite{simonyan2014very},  ResNet~\cite{he2016deep}, and MobileNet-V2~\cite{sandler2018mobilenetv2}
on CIFAR~\cite{krizhevsky2009learning} and ILSVRC-2012~\cite{deng2009imagenet}. 
We report compressed models at different FLOPs by adding a letter suffix (e.g., \method-A and \method-B).
We compare various state-of-the-art compression methods (including their baseline acc.), e.g.,
TAS~\cite{dong2019network},
LeGR~\cite{chin2020towards},
DMCP~\cite{guo2020dmcp},
SSR-GR~\cite{wang2021convolutional}, and CC~\cite{li2021towards},
where \method\ outperforms all of them.

\begin{table}[t]
\centering
\fontsize{8}{10}\selectfont

\begin{tabular}{|c|c|c|c|c|}
    \hline
    Network & Method & Test Acc. (\%) & Acc. $\downarrow$  & \thead{FLOPs (M) \\ Pruned (\%)}  \\
    \hline\hline
    

    \multirow{4}{3em}{VGG16} & L1 \cite{li2016pruning} & 93.25 $\rightarrow$ 93.40 & -0.15 & 211 (34.2) \\
    
    & GAL \cite{lin2019towards} & 93.96 $\rightarrow$ 93.42 & 0.54 & 172 (45.2) \\
    
    & SSS \cite{huang2018data} & 93.96 $\rightarrow$ 93.02 & 0.94 & 183 (39.6) \\

    & {\bf \method} & {\bf 93.45 $\rightarrow$ 93.78} & {\bf -0.33} & {\bf 62 (80.1)} \\ 

    \hline\hline
    
    
    \multirow{9}{3em}{\thead{ResNet \\ 56}} & GAL \cite{lin2019towards} & 93.26 $\rightarrow$ 93.38 & -0.12 & 78 (37.6) \\
    
    & NISP \cite{yu2018nisp} & 93.04 $\rightarrow$ 93.01 & 0.03 & 71 (43.6)  \\
    
    & DCP \cite{zhuang2018discrimination} & 93.80 $\rightarrow$ 93.49 & 0.31 &  63 (49.8)  \\
    
    & {\bf \method-A} & {\bf 93.60 $\rightarrow$ 94.01} & {\bf -0.41} & {\bf 63 (49.6)}
    \\ \cdashline{2-5}

    & TAS \cite{dong2019network} & 94.46 $\rightarrow$ 93.69 & 0.77 & 60 (52.0) \\
    
    & FPGM \cite{he2019filter} & 93.59 $\rightarrow$ 93.49 & 0.10 & 59 (52.6) \\
    
    & LFPC~\cite{he2020learning} & 93.59 $\rightarrow$ 93.24 & 0.35 & 59 (52.9) \\
    
    & KSE \cite{li2019exploiting} & 93.03 $\rightarrow$ 92.88 & 0.15 & 50 (60.0) \\    
    
    & {\bf \method-B} & {\bf 93.60 $\rightarrow$ 93.86} & {\bf -0.26} & {\bf 46 (63.0)} \\ 
    \hline\hline
    

    \multirow{7}{3em}{\thead{ResNet \\ 110}} & L1 \cite{li2016pruning} & 93.53 $\rightarrow$ 93.30 & 0.23 & 155 (38.6)   \\
    
    & NISP \cite{yu2018nisp} & 93.53 $\rightarrow$ 93.38 & 0.15 & 142 (43.8)  \\ 
    
    & {\bf \method-A} & {\bf 93.70 $\rightarrow$ 94.42} & {\bf -0.72} & {\bf 136 (46.1)}\\ 
    \cdashline{2-5}
    
    & GAL \cite{lin2019towards} & 93.50 $\rightarrow$ 92.74 & 0.76 & 130 (48.5)  \\
    
    & FPGM \cite{he2019filter} & 93.68 $\rightarrow$ 93.74 & -0.06 & 121 (52.3)\\
    
     & LFPC~\cite{he2020learning} & 93.68 $\rightarrow$ 93.07 & 0.61 & 101 (60.0) \\
    
    & {\bf \method-B} & {\bf 93.70 $\rightarrow$ 94.08} & {\bf -0.38} & {\bf 101 (60.0)}\\

    \hline
    \end{tabular}
	\vspace{-0.3cm}    
    \caption{CIFAR-10 Compression Results}\label{tab:CIFAR10_Experiments}

\smallskip

 \begin{tabular}{|c|c|c|c|c|}
    \hline
    Network & Method & Test Acc. (\%) & Acc. $\downarrow$  & \thead{FLOPs (M) \\ Pruned (\%)} \\
    \hline\hline
    
    
    \multirow{2}{3em}{VGG19} & SLIM~\cite{liu2017learning} & 73.26 $\rightarrow$ 73.48 & -0.22 & 256 (37.1) \\
    
    & {\bf \method} & {\bf 73.40 $\rightarrow$ 73.96} & {\bf -0.56} & {\bf 161 (59.5)} \\ 

    \hline\hline

       \multirow{5}{3em}{\thead{ResNet \\ 56}} 
    & SFP~\cite{he2018soft} & 71.33 $\rightarrow$ 68.37 & 2.96 & 76 (39.3) \\ 
    
    & FPGM~\cite{he2019filter} & 71.40 $\rightarrow$ 68.79 & 2.61 & 59 (52.6) \\ 
    
    & LFPC~\cite{he2020learning} & 71.33 $\rightarrow$ 70.83 & 0.58 & 61 (51.6) \\ 
    
    & LeGR~\cite{chin2020towards} & 72.41 $\rightarrow$ 71.04 & 1.37 & 61 (51.4) \\
    
    & \textbf{\method} & \textbf{72.10 $\rightarrow$ 72.35} & \textbf{-0.25} & \textbf{55 (56.2)} \\
    
    \hline\hline
    
    
    \multirow{4}{3em}{\thead{ResNet \\ 164}} & LCCL \cite{dong2017more} & 75.67 $\rightarrow$ 75.26 & 0.41 & 197 (21.3)\\
    
    & SLIM~\cite{liu2017learning} & 76.63 $\rightarrow$ 76.09 & 0.54 & 124 (50.6) \\
    
    & DI~\cite{kung2019methodical} & 77.63 $\rightarrow$ 76.11 & 1.52 & 105 (58.0)\\

    & {\bf \method} & {\bf 78.00 $\rightarrow$ 78.05} & {\bf -0.05} & {\bf 92 (63.2)}\\
    
    \hline
    \end{tabular}
    \vspace{-0.3cm}    
    \caption{CIFAR-100 Compression Results}\label{tab:CIFAR100_Experiments}
	
    \vspace{-0.3cm}

\end{table}

\begin{table*}[t]

\centering

\fontsize{8.5}{10}\selectfont

\begin{tabular}{|c|c|c|c|c|c|c|c|}
    \hline
    Network & Method & \thead{ Top-1 \\ Acc. (\%)}  & \thead{ Top-1 \\ $\downarrow$ (\%)} & \thead{Top-5 \\ Acc. (\%)} & \thead{Top-5 \\ $\downarrow$ (\%)} & \thead{FLOPs (B) \\ Pruned (\%)}  & \thead{Params (M) \\ Pruned (\%)} \\
    \hline\hline
    


    \multirow{10}{3.5em}{\thead{ResNet \\ 50}}    & GAL \cite{lin2019towards} & 76.15 $\rightarrow$ 71.95 & 4.20 & 92.87 $\rightarrow$ 90.94 & 1.97 & 2.33 (43.0) & 21.2 (16.9) \\
    
      & SFP \cite{he2018soft} & 76.15 $\rightarrow$ 74.61 & 1.54 & 92.87 $\rightarrow$ 92.06 & 0.81 & 2.38 (41.8) & - \\
    
    & HRank~\cite{lin2020hrank} & 76.15 $\rightarrow$ 74.98 & 1.17 & 92.87 $\rightarrow$ 92.33 & 0.54 & 2.30 (43.7) & 16.2 (36.5) \\
    
        &  SSR-GR \cite{wang2021convolutional} & 76.13 $\rightarrow$ 75.76 & 0.37 & 92.86 $\rightarrow$ 92.67 & 0.19 & 2.29 (44.1) & - \\
    
    & TAS~\cite{dong2019network} & 77.46 $\rightarrow$ 76.20 & 1.26 & 93.55 $\rightarrow$ 93.07 & 0.48 & 2.31 (43.5) & - \\
    
   & DMCP~\cite{guo2020dmcp} & 76.60 $\rightarrow$ 76.20 & 0.40 & - & - & 2.20 (46.0) & - \\
    
    & \textbf{\method-A} & \textbf{76.85 $\rightarrow$ 76.89} & \textbf{-0.04} & \textbf{93.17 $\rightarrow$ 93.33} & \textbf{-0.16} & \textbf{2.28 (44.3)} & \textbf{14.9 (41.6)} \\ 
    \cdashline{2-8}
    
    & FPGM~\cite{he2019filter} & 76.15 $\rightarrow$ 74.83 & 1.32 & 92.87 $\rightarrow$ 92.32 & 0.55 & 1.90 (53.5) & - \\
    
    & GBN~\cite{zhonghui2019gate} & 75.85 $\rightarrow$ 75.18 & 0.67 & 92.67 $\rightarrow$ 92.41 & 0.26 & 1.85 (55.0) & 11.9 (53.4) \\
    
    & LeGR~\cite{chin2020towards} & 76.10 $\rightarrow$ 75.30 & 0.80 & 92.90 $\rightarrow$ 92.40 & 0.50 & 1.93 (53.0) & - \\

    & \textbf{\method-B} & \textbf{76.85 $\rightarrow$ 76.35} & \textbf{0.50} & \textbf{93.17 $\rightarrow$ 93.04} & \textbf{0.13} & \textbf{1.90 (53.5)} & \textbf{12.7 (50.3)} \\ 
    


    \hline\hline

    
    \multirow{8}{4em}{\thead{ResNet \\ 18}}     & LCCL \cite{dong2017more} & 69.98 $\rightarrow$ 66.33 & 3.65 & 89.24 $\rightarrow$ 86.94 & 2.30 & 1.18 (34.6) & 11.7 (0.0) \\
    
    &  SLIM \cite{liu2017learning} & 68.98 $\rightarrow$ 67.21 & 1.77 & 88.68 $\rightarrow$ 87.39 & 1.29 & 1.31 (28.0) & - \\

     
     & \textbf{\method-A} & \textbf{70.05 $\rightarrow$ 69.39} & \textbf{0.66} & \textbf{89.40 $\rightarrow$ 88.81} & \textbf{0.59} & \textbf{1.15 (36.5)} & \textbf{7.3 (37.0)}
    \\\cdashline{2-8}         
    
    & SFP \cite{he2018soft} & 70.28 $\rightarrow$ 67.10  & 3.18 & 89.63 $\rightarrow$ 87.78 & 1.85 & 1.06 (41.8) & - \\

    & DCP \cite{zhuang2018discrimination} & 69.64 $\rightarrow$ 67.35 & 2.29 & 88.98 $\rightarrow$ 87.60 & 1.38 & 0.98 (46.0) & 6.2 (47.0) \\ 
    
    & FPGM \cite{he2019filter} & 70.28 $\rightarrow$ 68.41 & 1.87 & 89.63 $\rightarrow$ 88.48 & 1.15 & 1.06 (41.8) & - \\
    
     
     & \textbf{\method-B} & \textbf{70.05 $\rightarrow$ 68.86} & \textbf{1.19} & \textbf{89.40 $\rightarrow$ 88.61} & \textbf{0.79} & \textbf{1.05 (41.9)} & \textbf{6.7 (42.5)}
    \\
    
    \hline\hline
        
 \multirow{8}{4em}{\thead{MobileNet \\ V2}} &  Uniform~\cite{sandler2018mobilenetv2} & 71.80 $\rightarrow$ 69.80 & 2.00 & - & - & 0.22 (26.9) & - \\

& AMC~\cite{he2018amc} & 71.80 $\rightarrow$ 70.80 & 1.00 & - & - & 0.22 (26.9) & - \\

& CC~\cite{li2021towards} & 71.88 $\rightarrow$ 70.91 & 0.97 & - & - & 0.22 (28.3) & - \\

& Meta~\cite{liu2019metapruning} & 72.70 $\rightarrow$ 71.20 & 1.50 & - & - & 0.22 (27.9) & - \\

& \textbf{\method-A} & \textbf{72.18 $\rightarrow$ 71.97} & \textbf{0.21} & \textbf{90.49 $\rightarrow$ 90.39} & \textbf{0.10} & \textbf{0.22 (26.9)} & \textbf{2.8 (20.4)} \\
\cdashline{2-8}         

& DCP \cite{zhuang2018discrimination} & 70.11 $\rightarrow$ 64.22 & 5.89 & - & 3.77 & 0.17 (44.7) & 2.6 (25.9) \\

& Meta~\cite{liu2019metapruning} & 72.70 $\rightarrow$ 68.20 & 4.50 & - & - & 0.14 (53.4) & - \\

& \textbf{\method-B} & \textbf{72.18 $\rightarrow$ 69.22} & \textbf{1.96} &  \textbf{90.49 $\rightarrow$ 88.69} & \textbf{1.80} & \textbf{0.14 (53.4)} & \textbf{2.1 (39.3)} \\




\hline

\end{tabular}
\vspace{-0.3cm}
\caption{ILSVRC-2012 Compression Results}
\label{tab:ImageNet_Experiments}
\vspace{-0.35cm}

\end{table*}












\textbf{Training Settings.}
We use the Nesterov SGD optimizer~\cite{nesterov1983method} with a momentum of 0.9.
The weight decay factor is set to be 0.0001.
We use the standard data augmentation scheme~\cite{he2016deep} for all datasets. 
On CIFAR, we fine-tune 200 epochs with a minibatch size of 128. 
The learning rate is initialized at 0.05 and multiplied by 0.13 at epoch 80 and 160. 
On ILSVRC-2012, we use a batch size of 128 to fine-tune 100 epochs.
The learning rate is started at 0.025 with a cosine decay learning rate schedule.

\textbf{CDC Settings.} 
We adopt the same one-shot uniform hierarchical pruning scheme\footnote{For pruned MobileNet-V2, we round the number of channels to its closest integer that is divisible by 8 in each layer, as suggested in ~\cite{sandler2018mobilenetv2}.} as in Sec.~\ref{sec:hier_pruning_exp}.
On CIFAR, all training data are used for \metric\ channel scoring,
while we randomly sample 10,000 training images for \metric\ scoring on ILSVRC-2012.
We learn 5/20 coarse classes for CIFAR-10/100 and 
100 coarse classes for ILSVRC-2012 by spectral clustering on the confusion matrix.
We set the watershed layers for all networks to be the middle layers. 
For distillation,
we set $\lambda = 10.0$ and $\gamma = 1.0$ in Eqn.~\ref{eqn:training_loss} and 
learn the DCA weights by coarse class with the intermediate loss only inserted at the watershed layer.
The student DCA is updated at the 40\% and 80\% of the total epochs.

\textbf{CIFAR Results.}
On CIFAR-10, 
our VGG-16 has a 3$\times$ inference speedup with respect to GAL~\cite{lin2019towards},  
while having 0.36\% higher accuracy. 
Our \method-A of ResNet-56 outperforms DCP~\cite{zhuang2018discrimination}
with 0.52\% accuracy gain at the same FLOPs reduction ratio.
Compared to LFPC~\cite{he2020learning} on ResNet-110, 
\method-B achieves 94.08\% accuracy, which is 1.01\% higher with the same FLOPs. 
On CIFAR-100, our compressed VGG-19 has an accuracy gain of 0.48\% compared to SLIM~\cite{liu2017learning} with 22.4\% less FLOPs. 
Compared to LFPC~\cite{he2020learning} and LeGR~\cite{chin2020towards}, 
\method\ achieves
an accuracy gain of 1.52\% and 1.31\%, respectively, 
while having 5\% less FLOPs on ResNet-56.
On ResNet-164, 
\method\ outperforms DI~\cite{kung2019methodical} by 1.94\% in accuracy 
with 5\% less FLOPs.

\textbf{ILSVRC-2012 Results.}
On ResNet-50, \method-A 
increases top-1 and top-5 accuracies from the baseline by 0.04\% and 0.16\% with 44.3\% FLOPs pruned.
Compared to DMCP~\cite{guo2020dmcp} and TAS~\cite{dong2017learning},
\method-A has a 0.69\% top-1 accuracy gain.
Moreover, \method-B achieves a top-1 accuracy of 76.35\% with 53.5\% FLOPs reduction, 
surpassing all prior methods.
On ResNet-18, \method-A achieves a 3.06\% accuracy gain with a higher FLOPs reduction ratio compared to LCCL~\cite{dong2017more}.
\method-B demonstrates top-1 accuracy gains of 1.76\% and 1.51\%
with respect to SFP~\cite{he2018soft} and DCP~\cite{zhuang2018discrimination}.
We finally evaluate \method\ on MobileNet-V2, which is a much compact network suitable for mobile deployment.
\method-A achieves a top-1 accuracy of 71.97\% with 26.9\% FLOPs reduction, outperforming AMC~\cite{he2018amc},  LeGR~\cite{chin2020towards}, and CC~\cite{li2021towards}.
\method-B advances DCP~\cite{zhuang2018discrimination} and Meta~\cite{liu2019metapruning} by 5\% and 1.02\% top-1 accuracy, when 53.4\% of FLOPs are pruned.
\method's superior performance well indicates the effectiveness of discriminative compression.

\section{Ablation Study} 

\begin{figure*}[t]
	\centering
    \includegraphics[width = 0.8\textwidth]{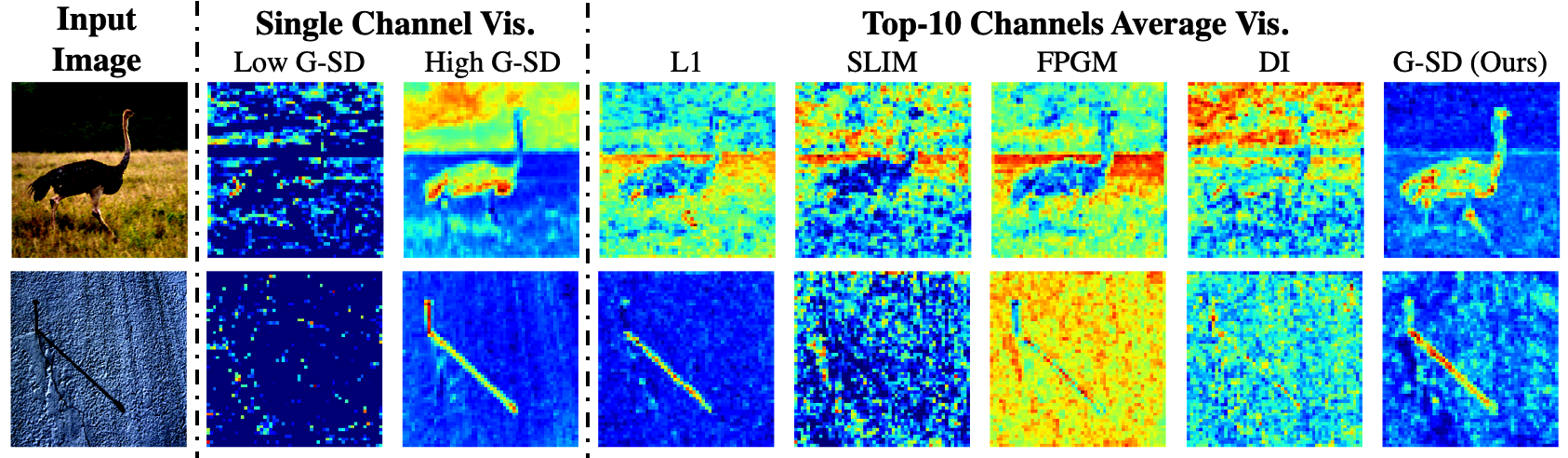}
    \vspace{-0.3cm}
    \caption{Channel selection analysis. 
    {\bf Col. 1:} Input images. 
    {\bf Col. 2-3}: Channels with low and high \metric\ values. The low \metric\ channel generates responses that are indistinguishable to different classes,
    while the high one produces informative activations.
    {\bf Col. 4-8}: Average responses of the top-10 channels selected by different metrics. 
    From left to right, the metrics are: $\ell$1-weight~\cite{li2016pruning}, batch-norm scaling factor~\cite{liu2017learning}, filter's geometric median~\cite{he2019filter}, DI~\cite{kung2019methodical}, and \metric.
    In general, the channels picked by \metric\ preserves the most classification information.
    }
    \label{fig:Qualitative Channel Selection}
    \vspace{-0.35cm}
\end{figure*}

\textbf{\metric.}
In Fig.~\ref{fig:Qualitative Channel Selection}, we visualize channels at \texttt{Res1\_2} in ResNet-50 on ILSVRC-2012 to intuitively show the effectiveness of \metric.
In \textbf{Col.~1-3}, 
we observe that the channel with low \metric\ ({\bf Col.~2}) tends to generate indistinguishable responses for different classes,
while the high one ({\bf Col.~3}) well preserves the informative image patterns  for classification. 
We further compare \metric\ with other channel selection criteria~\cite{li2016pruning,liu2017learning,he2019filter,kung2019methodical},
where we compute an average response over the top-10 highest scored channels for each metric, shown in \textbf{Col.~4-8}. 
We observe the average responses of \metric\ ({\bf Col.~8}) 
tends to display more class information than the others ({\bf Col.~4-7}).
In the first row, \metric\ clearly separates the ostrich from the background grass, while others generate mixed responses.
Moreover, \metric\ is the only one that preserves both the vertical nail and its long diagonal shadow in the second row.
Such visual comparisons demonstrate \metric's advantages over other metrics.

\begin{figure}[t]
    \centering
    \includegraphics[width=0.43\textwidth]{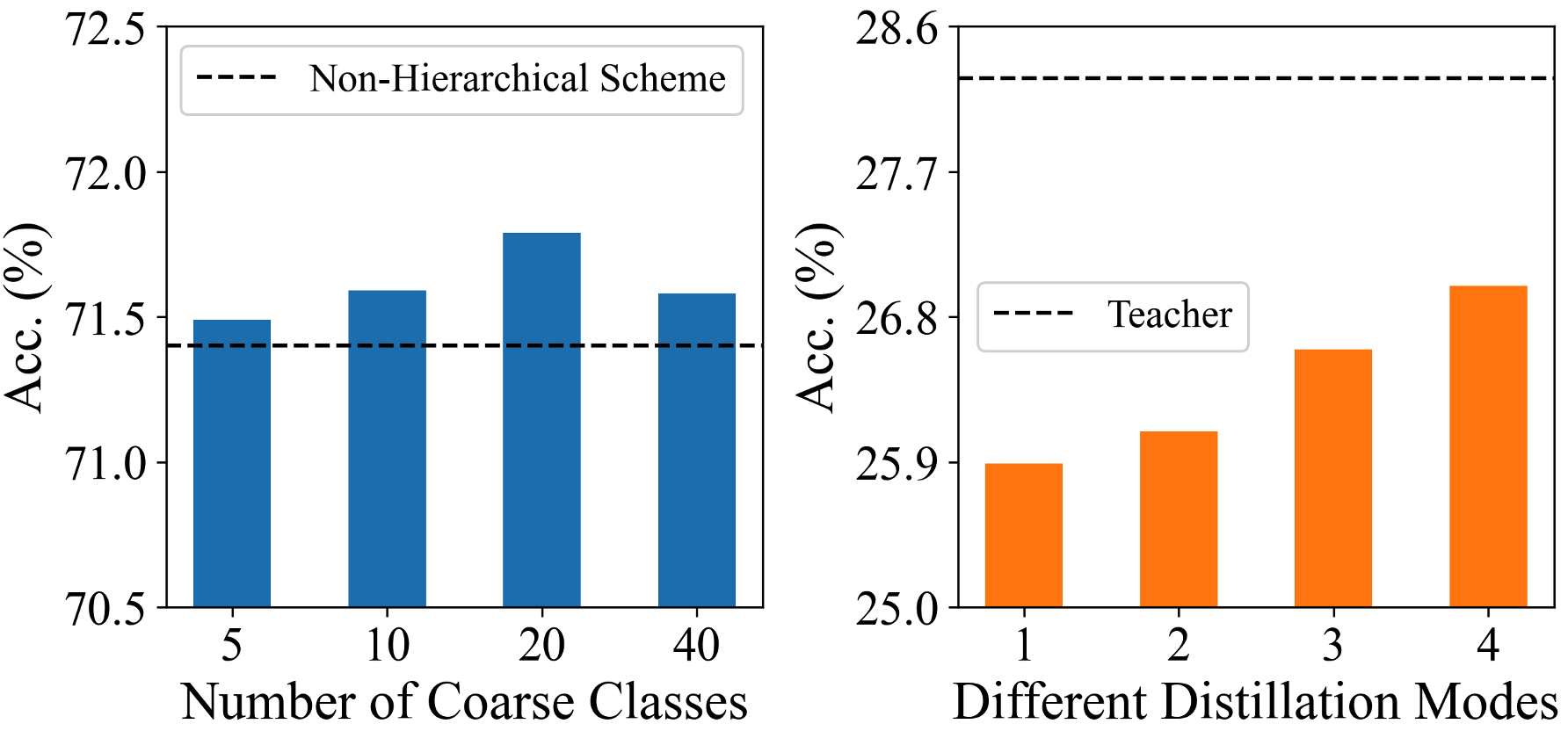}
    \vspace{-0.3cm}
    \caption{
    \textbf{Left:} Results of different number of coarse classes for hierarchical pruning. 
    Including an extra coarse class hierarchy improves network pruning performance,
    while using 20 coarse classes (same number of classes as ground truth) achieves the best performance.
    \textbf{Right:} Linear separability (measured by classification accuracy of linear SVM) of the intermediate layers trained by different intermediate distillation methods. 
    The layer trained by coarse class DCA distillation shows the best linear separability.
    }
    \label{fig:hp_dca_ablation_study}
    \vspace{-0.47cm}
\end{figure}

\textbf{Hierarchical Pruning.}
As the optimal number of clusters for spectral clustering is mainly based on different heuristics~\cite{lihi2004self,von2007tutorial},
we conduct an empirical study on the number of coarse classes we should learn for $\mathcal{Y}_c$ in hierarchical pruning.
We investigate it on CIFAR-100 and use $\mathcal{Y}_c$ of different number of coarse classes to hierarchically prune 56\% FLOPs from a ResNet-56 and retrain it only by $\mathcal{L}_{CE}$, 
with the results shown in Fig.~\ref{fig:hp_dca_ablation_study}. 
We find that using the coarse class information from $\mathcal{Y}_c$ generally improves over the non-hierarchical pruning.
Moreover, we find that learning a 20-class $\mathcal{Y}_c$  achieves the best result, 
which could be explained that the ground truth labeling of CIFAR-100 also has 20 coarse classes.

\textbf{DCA-Based Distillation.}
We measure the linear separability of the intermediate layers by evaluating their classification accuracies on linear SVMs.
We compare four different intermediate distillation schemes:
(1) No distillation. (2) Hint layer~\cite{romero2014fitnets}. (3) Fine class DCA. 
(4) Coarse class DCA.
We train linear SVMs (SVM parameter C searched on [1e-5, 1e-4, ..., 1, 10]) on the watershed layers of ResNet-164 (both full-size teacher and 63\%-pruned students) on CIFAR-100 with results in Fig.~\ref{fig:hp_dca_ablation_study}.
We find that the hint layer~\cite{romero2014fitnets} (\textbf{Mode 2}) has only slight improvement on layer's linear separability compared to no distillation (\textbf{Mode 1}).
Moreover, the coarse class DCA scheme (\textbf{Mode 4}) best improves the linear separability of the student's hidden layer.

\section{Conclusion} 

In this paper, 
we propose class-discriminative compression (\method) to learn efficient neural networks. 
While limited attempts has been made to leverage classification information for network compression, 
we design a unified framework for discriminative pruning and distillation,
fitting seamlessly with the discriminative training objective.
To better identify channels' redundancy for class-discriminative pruning,
we study the pruning effectiveness of a group of closed-form discriminant functions 
and propose a hierarchical pruning paradigm.
Moreover, we make the first attempt to distill discriminative information in hidden layers' subspace by discriminant component analysis.
Combing the pruning and distillation approaches, 
\method\ is evaluated on CIFAR and ILSVRC-2012,
outperforming state-of-the-art methods by a clear margin.

\bibliographystyle{Style_Ref/ieee_fullname}
\bibliography{Style_Ref/refs}

\newpage

The supplementary material is organized as follows. 
In Sec.~\ref{sec:DF_study}, we present the definitions of our group of discriminant functions and a mathematical analysis on our adopted metric, \metric.  
The detailed settings of the experiments in our main paper are included in Sec.~\ref{sec:study_details}.
We discuss 7 additional ablation studies on \metric, 
hierarchical pruning,
and DCA-based distillation in Sec.~\ref{sec:extra_ablation}.
Additional visualizations of 
the channel selection analysis are provided in Sec.~\ref{sec:visual}.
\section{Discriminant Functions}\label{sec:DF_study}

\subsection{Single-Variate Binary-Class Metrics}

We adopt the same notation of the $n$-sample binary-class single-variate dataset as in the main paper.
Our implementations of Symmetric Divergence (SD)~\cite{mak2006solution}, Absolute SNR (AbsSNR)~\cite{golub1999molecular},
Fisher Discriminant Ratio (FDR)~\cite{pavlidis2001gene}, and Student's T-Test (Ttest)~\cite{lehmann2006testing} are:
\begin{equation}\label{eqn:SD}
   \mathrm{SD}(\mathcal{D}, \mathcal{B}) = \frac{1}{2} \left( 
\frac{\sigma_{\mathcal{D}^+}^2}{\sigma_{\mathcal{D}^-}^2} + \frac{\sigma_{\mathcal{D}^-}^2}{\sigma_{\mathcal{D}^+}^2}
\right) 
+ 
\frac{1}{2}
\left( 
\frac{ (\mu_{\mathcal{D}^+} - \mu_{\mathcal{D}^-})^2 }
{\sigma_{\mathcal{D}^+}^2 + \sigma_{\mathcal{D}^-}^2}
 \right) - 1 
\end{equation}
\begin{equation}
\mathrm{AbsSNR}(\mathcal{D}, \mathcal{B}) = \frac{|\mu_{\mathcal{D}^+} - \mu_{\mathcal{D}^-}|}
{\sigma_{\mathcal{D}^+} + \sigma_{\mathcal{D}^-}}
\end{equation}
\begin{equation}
\mathrm{FDR}(\mathcal{D}, \mathcal{B})  = \frac{(\mu_{\mathcal{D}^+} - \mu_{\mathcal{D}^-})^2}
{\sigma_{\mathcal{D}^+}^2 + \sigma_{\mathcal{D}^-}^2}
\end{equation}
\begin{equation}\label{eqn:Ttest}
\mathrm{Ttest}(\mathcal{D}, \mathcal{B})= \frac{|\mu_{\mathcal{D}^+} - \mu_{\mathcal{D}^-}|}
{\sqrt{
\frac{\sigma_{\mathcal{D}^+}^2}{|\mathcal{D}^+|}
+ \frac{\sigma_{\mathcal{D}^-}^2}{|\mathcal{D}^-|}
}
}
\end{equation}
where $|\mathcal{D}^+|$ and $|\mathcal{D}^-|$ denote the number of samples in the two classes of $\mathcal{D}$.

We observe that the computation of the three metrics (not appeared in the main paper), AbsSNR, FDR, and Ttest, mainly requires four statistics, $\mu_{\mathcal{D}^+}$, $\sigma_{\mathcal{D}^+}^2$, $\mu_{\mathcal{D}^-}$, and $\sigma_{\mathcal{D}^-}^2$, which are the same as SD.
Therefore, we can adopt the same way explained in the main paper to generalize them for channel scoring:
(1) Partition the feature maps of a channel in a one-versus-all manner based on their labels;
(2) Apply specialized statistics extractors to obtain the metric score for each partition;
(3) Aggregate the scores from all partitions to a single scalar.
Via such way, we derive G-AbsSNR, G-FDR, and G-Ttest for high-dimensional multi-label channel scoring and pruning.

We also provide an analysis on the mathematical merits of SD as follows.
The discriminant power of a scalar feature is quantified by the first two terms in SD.
The first term measures the {\itshape divergence} of two classes {\itshape symmetrically}. 
It achieves a high score for the feature whose $\sigma_{\mathcal{D}^+}^2$ or $\sigma_{\mathcal{D}^-}^2$ is much smaller than the other,
which indicates that one class of the feature has a more concentrated distribution, and thus the feature is more class distinguishable. 
The second term\footnote{The expressions of SD's second term vary a bit in~\cite{kung2014kernel} and~\cite{mak2006solution}.
In this work, we adopt the one in~\cite{kung2014kernel}.
} 
awards features that have large ``Signal-to-Noise Ratio (SNR)'' \cite{jaynes1957information} 
where the centroid distance $(\mu_{\mathcal{D}^+} - \mu_{\mathcal{D}^-})^2$ can be viewed as ``class discrimination signal" and the class variance sum $\sigma_{\mathcal{D}^+}^2 + \sigma_{\mathcal{D}^-}^2$ is regarded as ``class variational noise''.
Therefore, SD's ability to find discriminant features is mathematically supported.

\subsection{High-Dimensional Metrics}

We present the definitions of Discriminant Information (DI)~\cite{kung2019methodical} and Maximum Mean Discrepancy (MMD)~\cite{gretton2012kernel} here.
We use the same notation of $N$-sample $Y$-class dataset in the main paper and regard $f_i$ as a flattened 1D vector.

\textbf{DI.} As each $f_i$ is associated with a label $y_i$, we can derive the scatter matrices $\mathbf{\bar{S}}$ and $\mathbf{S_B}$ for the feature map collection $\mathcal{F} = \{(f_i, y_i)\}_{i=1}^{N}$ via the same way as in the main paper. DI is then defined as:
\begin{equation}
\mathrm{DI}(\mathcal{F}) = tr([\mathbf{\bar{S}} + \rho\mathbf{I}]^{-1} \mathbf{S_B})
\end{equation}
where $\rho$ is a small ridge factor to ensure invertibility. 
In practice, we set $\rho = 0.0001$.

\textbf {MMD.} We define the rbf kernel as:
\begin{equation}
k(x,y) = exp\{-\frac{
||x - y ||^2_2}
{2 \sigma^2}\}
\end{equation}
The two-class MMD on a $(\feamap^c, \feamap^{-c})$ partition can thus be defined as:
\begin{align}
\nonumber\mathrm{MMD}_{two}(\feamap^c, \feamap^{-c}) &= \frac{1}{|\feamap^c|^2}\sum_{f_i,f_j\in\feamap^c}{k(f_i, f_j)} 
\\ &\nonumber+\frac{1}{|\feamap^{-c}|^2}\sum_{f_i,f_j\in\feamap^{-c}}{k(f_i, f_j)} \\ &-\frac{2}{|\feamap^{c}|\times|\feamap^{-c}|}\sum_{f_i\in\feamap^c, f_j\in\feamap^{-c}}{k(f_i, f_j)} 
\end{align}
following the similar one-versus-all setting, the MMD score of a channel is given as:
\begin{equation}
\mathrm{MMD}(\feamap) = \frac{1}{Y}\sum_{c = 1}^Y{\mathrm{MMD}_{two}(\feamap^c , \feamap^{-c})}
\end{equation}
In practice, we set $\sigma = 1$ for the rbf kernel.

\section{Experimental Details}\label{sec:study_details}

\subsection{Training Parameters}
For the experiments we conduct in Sec.~4.1-4.3 and Sec.~5 of the main paper, we use the following settings to retrain our pruned networks.

We use the SGD optimizer with Nesterov momentum~\cite{nesterov1983method} to fine-tune VGG-16 and ResNet-38/56/164, where the momentum is set as 0.9.
The fine-tuning process takes 200 epochs with a batch size of 128, and the weight decay is set to be 1e-3.
We augment the training samples with a standard data augmentation scheme~\cite{he2016deep}.
The learning rates of VGG-16 and ResNet-38/56/164 are started as 0.006 and 0.05, 
and multiplied by 0.28 and 0.14 at 40\% and 80\% of the total number of epochs.
Our training codes are implemented in PyTorch~\cite{paszke2017automatic}.

\subsection{CDC Parameters}

To learn the coarse class hierarchy, 
we use scikit-learn~\cite{scikit-learn} to implement the KMeans clustering\footnote{\url{https://scikit-learn.org/stable/modules/generated/sklearn.cluster.KMeans.html}} as well as the spectral clustering\footnote{\url{https://scikit-learn.org/stable/modules/generated/sklearn.cluster.SpectralClustering.html}} algorithm. 
The confusion matrices and the last hidden-layer activations are obtained from the full size ResNet-110, ResNet-164, and ResNet-50 for CIFAR-10, CIFAR-100, and ILSVRC-2012.
For our DCA-based distillation, 
we use Scipy~\cite{2020SciPy-NMeth} to solve Eqn.~6 for the DCA weights\footnote{\url{https://docs.scipy.org/doc/scipy/reference/generated/scipy.linalg.eigh.html}} and use scikit-learn to train linear SVMs\footnote{\url{https://scikit-learn.org/stable/modules/generated/sklearn.svm.LinearSVC.html}}. 
For coarse class DCA distillation, we use the coarse class learned by spectral clustering.
For the output knowledge distillation~\cite{hinton2015distilling}, we set the temperature parameter $T = 1$ for all experiments.

\section{Additional Ablation Study}\label{sec:extra_ablation}

\begin{figure}[t]
    \centering
    \includegraphics[width=0.35\textwidth, trim={0cm 0 0.1cm 0}, clip]{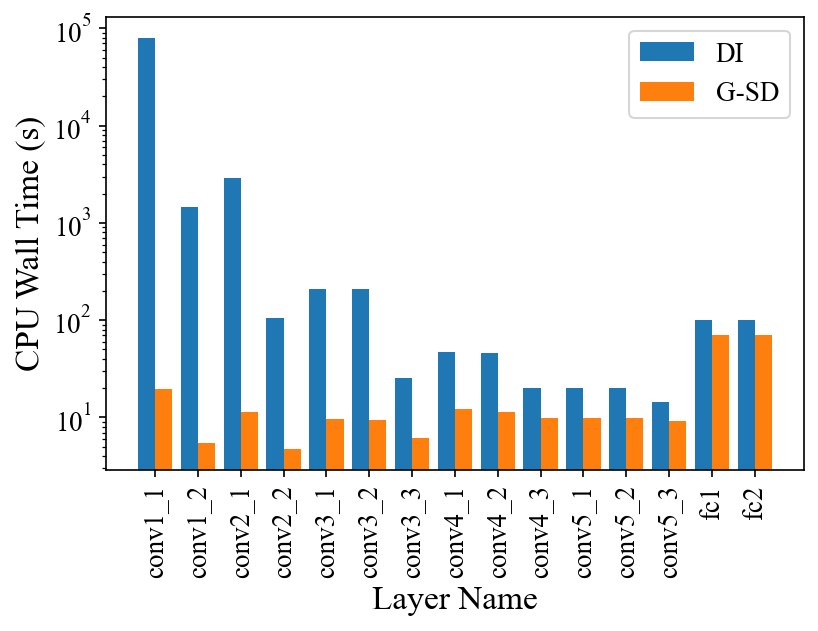}
    \caption{
        The CPU wall time for scoring all channels in VGG-16 layers by \metric\ and DI on ILSVRC-2012. 
        \metric\ shows a significant advantage in computational efficiency.
    }\label{fig:time_complexity}
    \vspace{-0.3cm}
\end{figure}
\begin{figure}[t]
        \centering
        \includegraphics[width=0.35\textwidth]{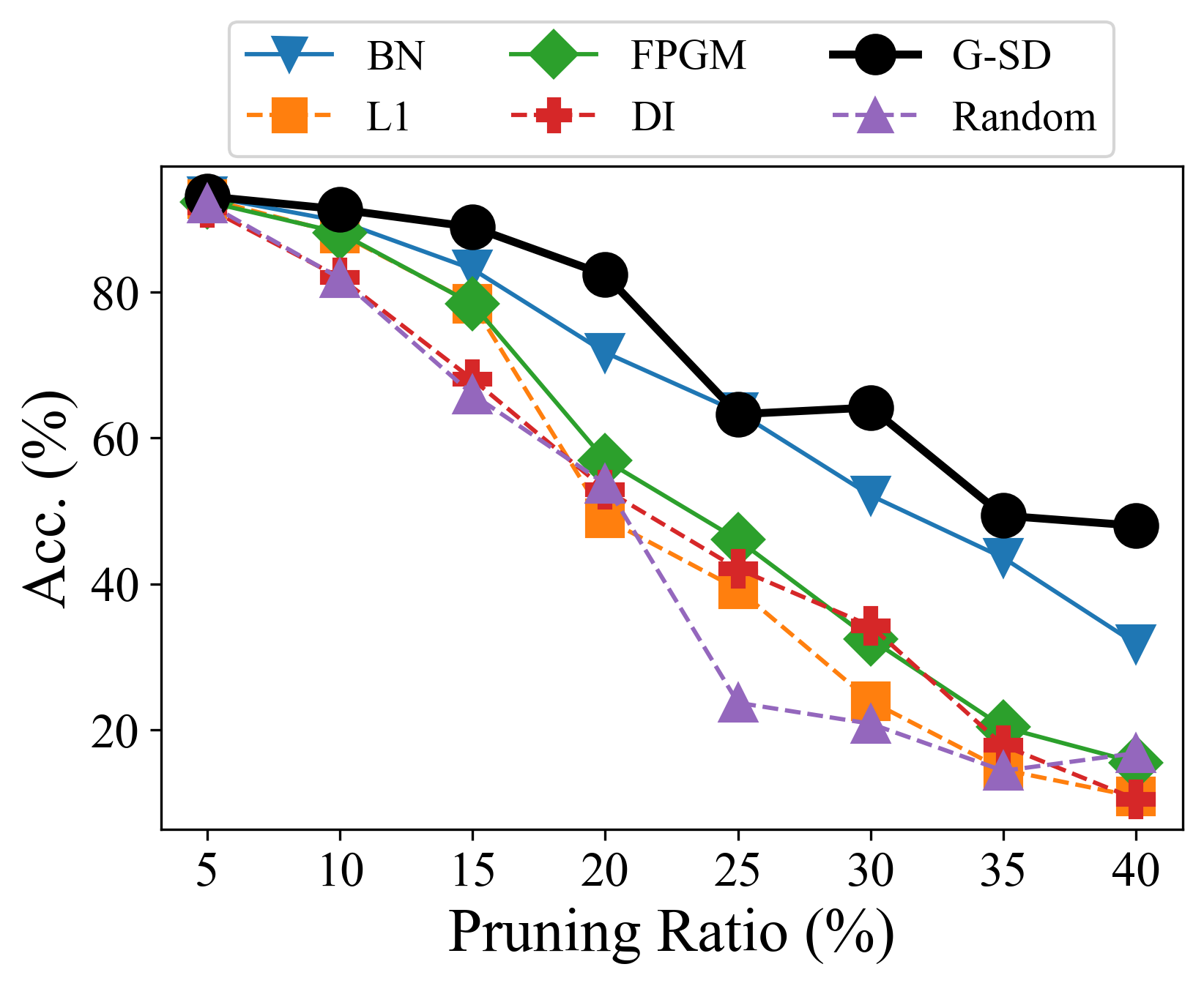}
        \caption{
        Layer pruning ratio vs. accuracy (without retraining) by \metric\ and other state-of-the-art pruning criteria with ResNet-110 on CIFAR-10. 
        \metric\ outperforms all other criteria by a clear margin. 
        }
        \label{fig:Quantitative Channel Selection}
        \vspace{-0.3cm}
\end{figure}

\subsection{Time Complexity of \metric.}

We conduct an empirical study to investigate the time complexity of two discriminant functions, \metric\ and DI, for channel scoring in Fig.~\ref{fig:time_complexity}.
We pick 3,000 ILSVRC-2012 samples and compute their feature maps in the channels of a VGG-16.
The maps are computed via an NVIDIA Tesla P100 GPU 
and the scorings of feature maps are executed on an Intel Xeon E5-2680 v4 CPU.
We calculate the CPU wall time for scoring all the channels in each layer 
by the two discriminant functions (excluding the time to obtain the maps).
We note that \metric\ generally has much less time complexity than DI in the scoring process of each layer.
Moreover, for layer $\mathrm{conv1\_1 }$ whose feature maps are of size 224$\times$224, \metric\ shows a 4000$\times$ speedup over DI.

\begin{figure*}[t]
    \centering
    \begin{subfigure}[t]{0.3\textwidth}
        \centering
        \includegraphics[width=\textwidth]{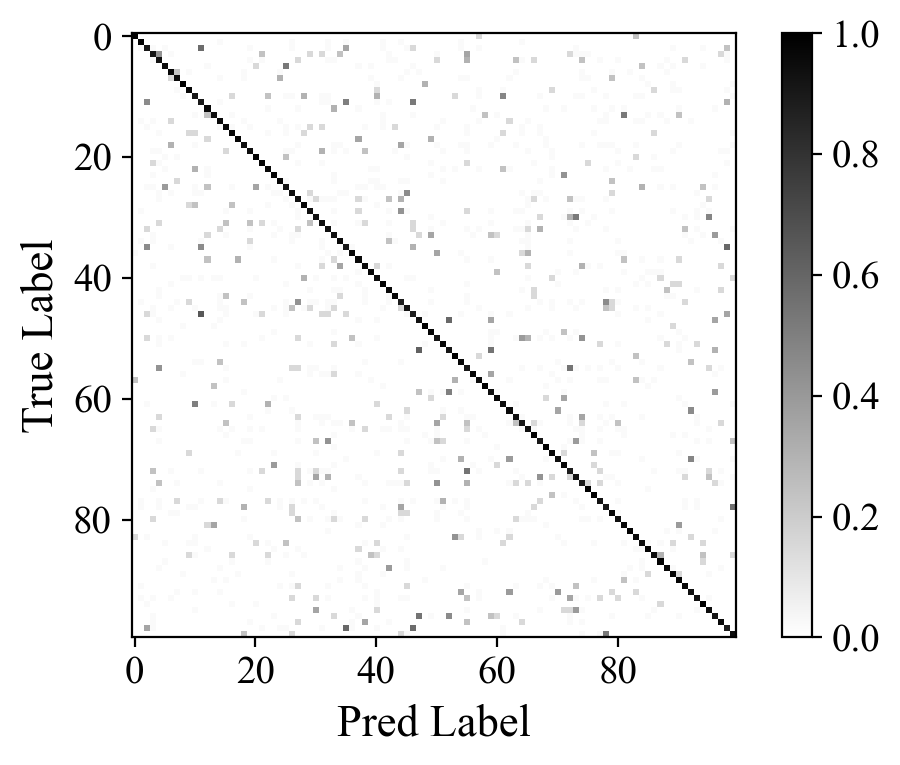}
        \caption{Ungrouped Confusion Matrix}\label{fig:ungrouped_class_hierarchy}
    \end{subfigure}%
    \hfill
    \begin{subfigure}[t]{0.3\textwidth}
        \centering
        \includegraphics[width=\textwidth]{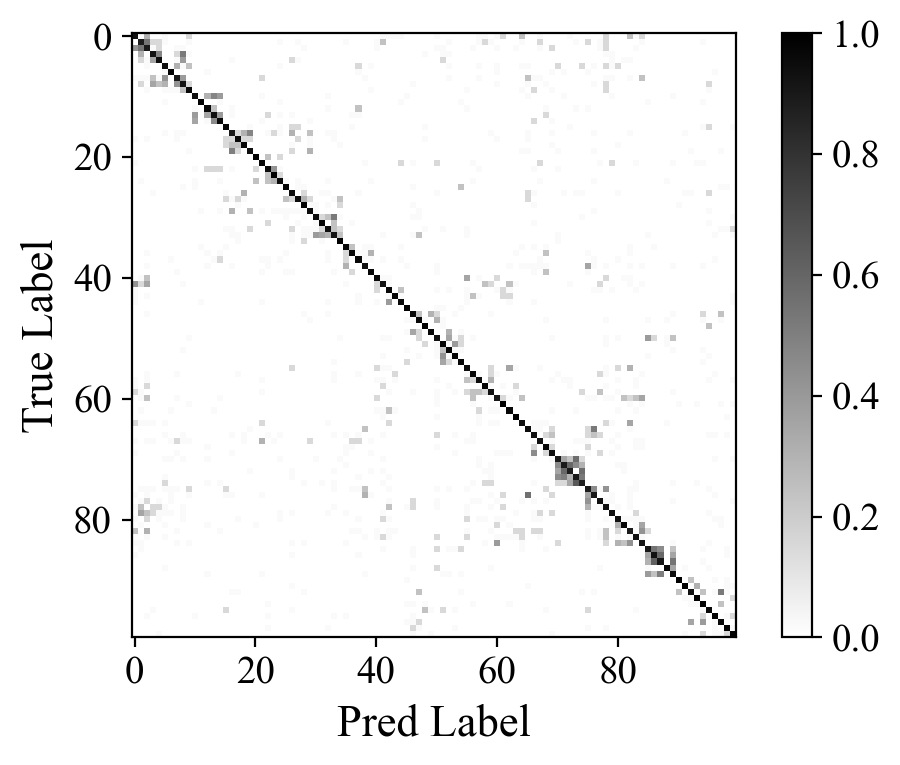}    
        \caption{Regrouped by Ground Truth}\label{fig:grouped_gt_class_hierarchy}    
    \end{subfigure}
    \hfill
     \begin{subfigure}[t]{0.3\textwidth}
        \centering
        \includegraphics[width=\textwidth]{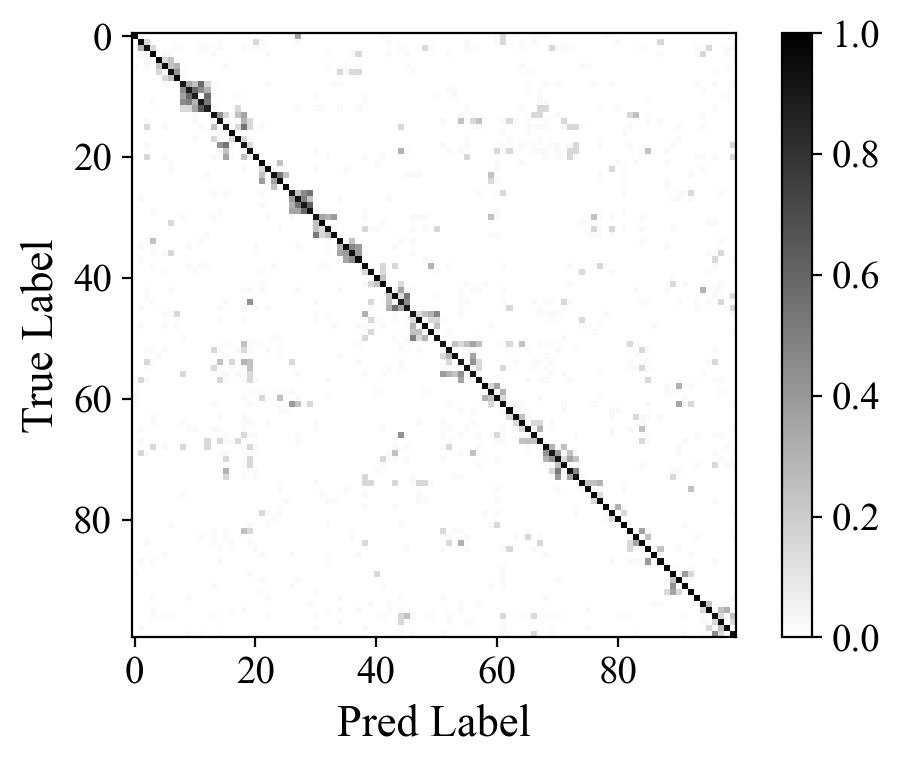}
        \caption{Regrouped by Learned Hierarchy}\label{fig:grouped_spec_class_hierarchy}
    \end{subfigure}
    \caption{Normalized CIFAR-100 confusion matrices. 
    (a) represents the confusion matrix with the original numbering scheme, 
    while (b) and (c) plot the confusion matrices regrouped by the coarse class hierarchies of ground truth and spectral-clustering learning.
    We observe block structures in both (b) and (c) where one class is easily predicted as the others within the same block.
    This indicates that the fine classes are similar to each other in the same block, sharing the same coarse semantics.
    Visually speaking, our spectral clustering method can explore the similarities among fine classes as well as the ground truth one.
    }
    \label{fig:CIFAR100 Confusion}
\end{figure*}

\subsection{Quantitative Channel Selection}

We demonstrate the advantage of \metric\ over other state-of-the-art pruning criteria~\cite{li2016pruning,liu2017learning,he2019filter,kung2019methodical} quantitatively in Fig.~\ref{fig:Quantitative Channel Selection} with ResNet-110 on CIFAR-10.
We use each metric to uniformly remove $r\%$ of the lowest-scored channels in all layers 
and include random scoring as the baseline, where $r \in \{5, 10, 15, 20, 25, 30, 35, 40\}$.
We then evaluate these pruned models' accuracies directly without retraining.
We observe that \metric\ wins against all other methods in nearly all FLOPs reduction ratios.

\subsection{Learned Class Hierarchy}

We further investigate the coarse hierarchy learned by spectral clustering on the confusion matrix.

On CIFAR-10, we learn 5 coarse classes, with the fine class grouping as:
(1) automobile, truck;
(2) airplane, ship;
(3) cat, dog;
(4) deer, horse;
(5) bird, frog.
These coarse classes can be labeled as vehicle transportation, other transportation tools, small mammals, large mammals, and non-mammals. 
This result shows that our method can well learn the coarse class semantics. 

On CIFAR-100, we show the confusion matrices of ResNet-164 in Fig.~\ref{fig:CIFAR100 Confusion}.
We first show the confusion matrix with the original label numbering scheme in Fig.~\ref{fig:ungrouped_class_hierarchy}.
Then, given a coarse class hierarchy, we rearrange the fine label number such that the fine classes in the same coarse class have adjacent label numbers 
(e.g., after grouping, fine label 1-5 corresponds to coarse label 1, etc.). 
We show the grouped confusion matrix by ground truth coarse hierarchy as well as the one learned by spectral clustering in Fig.~\ref{fig:grouped_gt_class_hierarchy} and~\ref{fig:grouped_spec_class_hierarchy}.
We find that our spectral clustering method can well explore the similarity among the fine classes, 
which is effective for learning the coarser class semantics.

\subsection{Hierarchical Pruning on ILSVRC-2012}

\begin{table}[t]
\fontsize{8.5}{10}\selectfont

\centering

\begin{tabular}{|c|c|c|c|c|}
\hline
Method & HRank~\cite{lin2020hrank} & FPGM~\cite{he2019filter} & LeGR~\cite{chin2020towards} &  \textbf{HP} \\ 
\hline
Param. (M) & 16.2 & - & - & \textbf{12.7} \\
 \hline
 FLOPs (B) & 2.30 & 1.90 & 1.93 & \textbf{1.90} \\
 \hline
 Acc. (\%) & 74.98 & 74.83 & 75.30 & \textbf{75.85} \\
\hline
\end{tabular}
\vspace{-0.2cm}
\caption{HP vs. SOTAs on ResNet-50 with ImageNet.}
\label{tab:HP_ResNet-50}

\medskip

\begin{tabular}{|c|c|c|c|c|}
\hline
Method & DCP~\cite{zhuang2018discrimination} & Uniform~\cite{sandler2018mobilenetv2} & Meta~\cite{liu2019metapruning} &  \textbf{HP} \\ 
\hline
Param. (M) & 2.6 & 2.6 & - & \textbf{2.1} \\
 \hline
 FLOPs (B) & 0.17 & 0.16 & 0.14 & \textbf{0.14} \\
 \hline
 Acc. (\%) & 64.22 & 68.25 & 68.20 & \textbf{68.76} \\
\hline
\end{tabular}
\vspace{-0.2cm}
\caption{HP vs. SOTAs on MobileNet-V2 with ImageNet.}
\label{tab:HP_MBV2}

\vspace{-0.2cm}

\end{table}

We further compare our hierarchical pruning (HP) scheme with state-of-the-art pruning methods on ILSVRC-2012 in Tab.~\ref{tab:HP_ResNet-50} and~\ref{tab:HP_MBV2}.
On both ResNet-50 and MobileNet-V2, 
HP again demonstrates its superiority by achieving higher accuracy with less inference cost.

\subsection{DCA-Based Distillation Weight}

\begin{figure}
    \centering
    \includegraphics[width=0.40\textwidth]{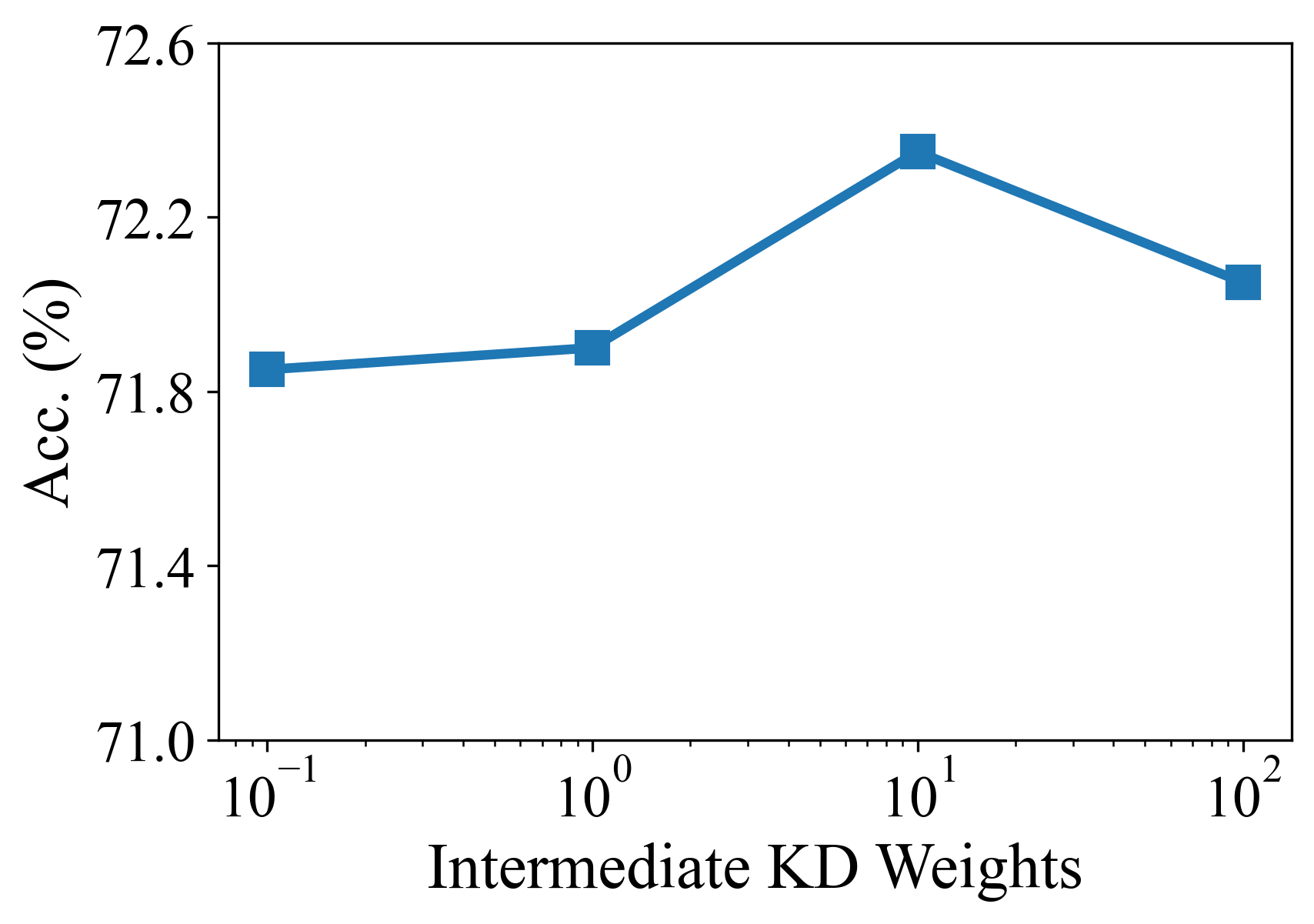}
    \caption{The accuracy curve for pruned ResNet-56 on CIFAR-100 varying the weights for DCA-based intermediate distillation. $\lambda = 10$ achieves the best result.}
    \label{fig:inter_kd_lambda}
\end{figure}

We investigate the knowledge distillation weight $\lambda$ for DCA-based intermediate distillation in Fig.~\ref{fig:inter_kd_lambda}. 
We use hierarchical pruning to prune 56\% FLOPs from a ResNet-56 trained on CIFAR-100,
and fine-tune it with output distillation weight $\gamma = 1.0$ while vary the intermediate distillation weight to be $\lambda \in \{0.1, 1, 10, 100\}$.
We find that setting $\lambda = 10$ achieves the best result for our scheme.

\subsection{PCA-Based Distillation}

\begin{figure}[t]
    \centering
    \includegraphics[width=0.40\textwidth]{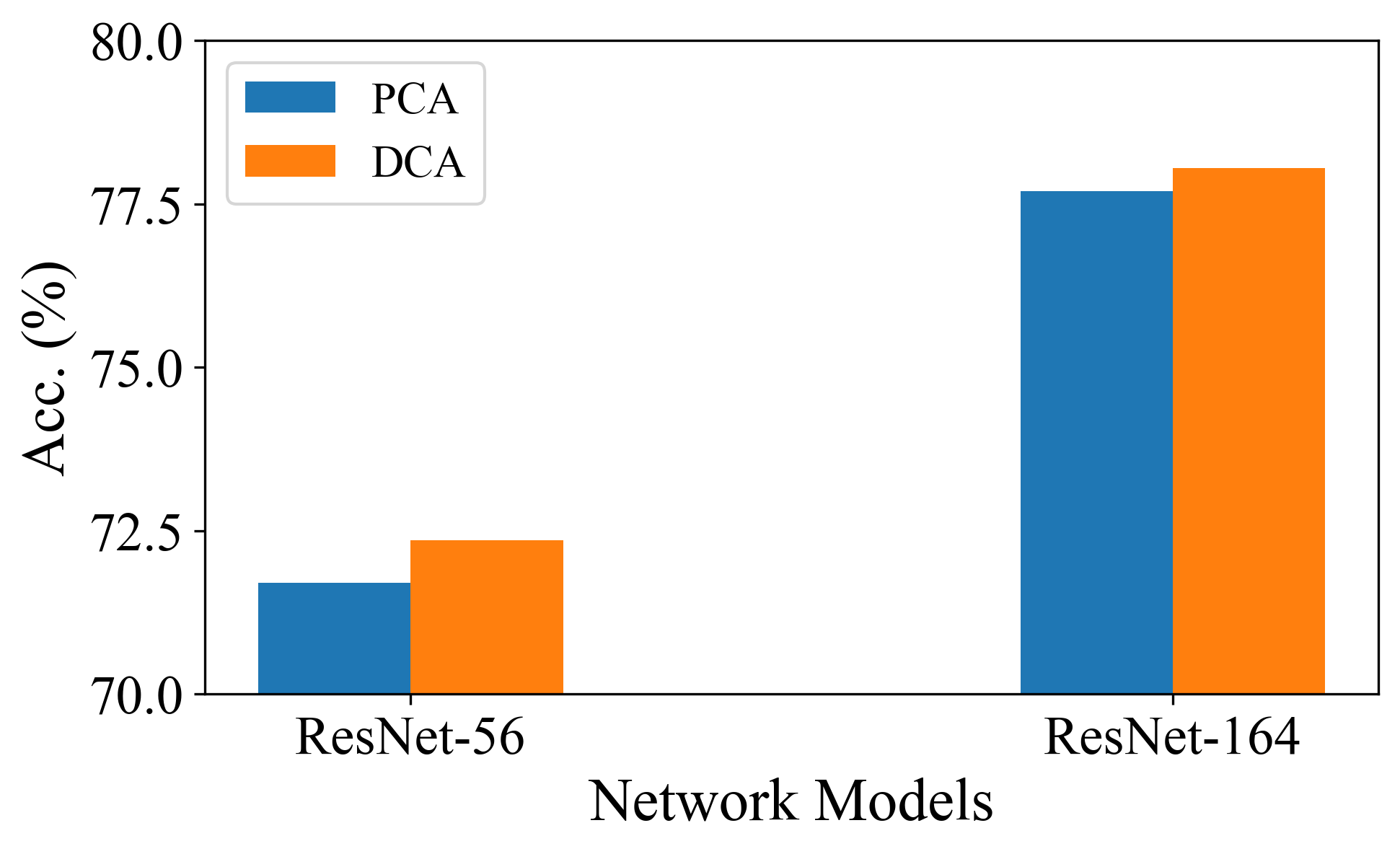}
    \caption{Comparing PCA with DCA for subspace distillation with ResNet-56 and ResNet-164 on CIFAR-100. 
    DCA outperforms PCA for the distillation quality on both networks. 
    }
    \label{fig:pca_vs_dca}
\end{figure}

We investigate using PCA components to transform layer activations for subspace distillation.
We keep all other settings (layer pruning ratio, weights learning frequency, inserted layers, etc.) to be the same and compare this scheme with our DCA-based distillation with pruned ResNet-56/164 (45\% of channels uniformly removed from each layer) on CIFAR-100 in Fig.~\ref{fig:pca_vs_dca}. 
We find that our DCA-based distillation generally improves over the PCA approach,
which agrees with the finding in the original paper~\cite{kung2017discriminant} that the subspace found by DCA is more class-discriminative than the PCA one.

\subsection{Multi-Hierarchy CDC}

\begin{figure}[t]
    \centering
    \includegraphics[width=0.40\textwidth]{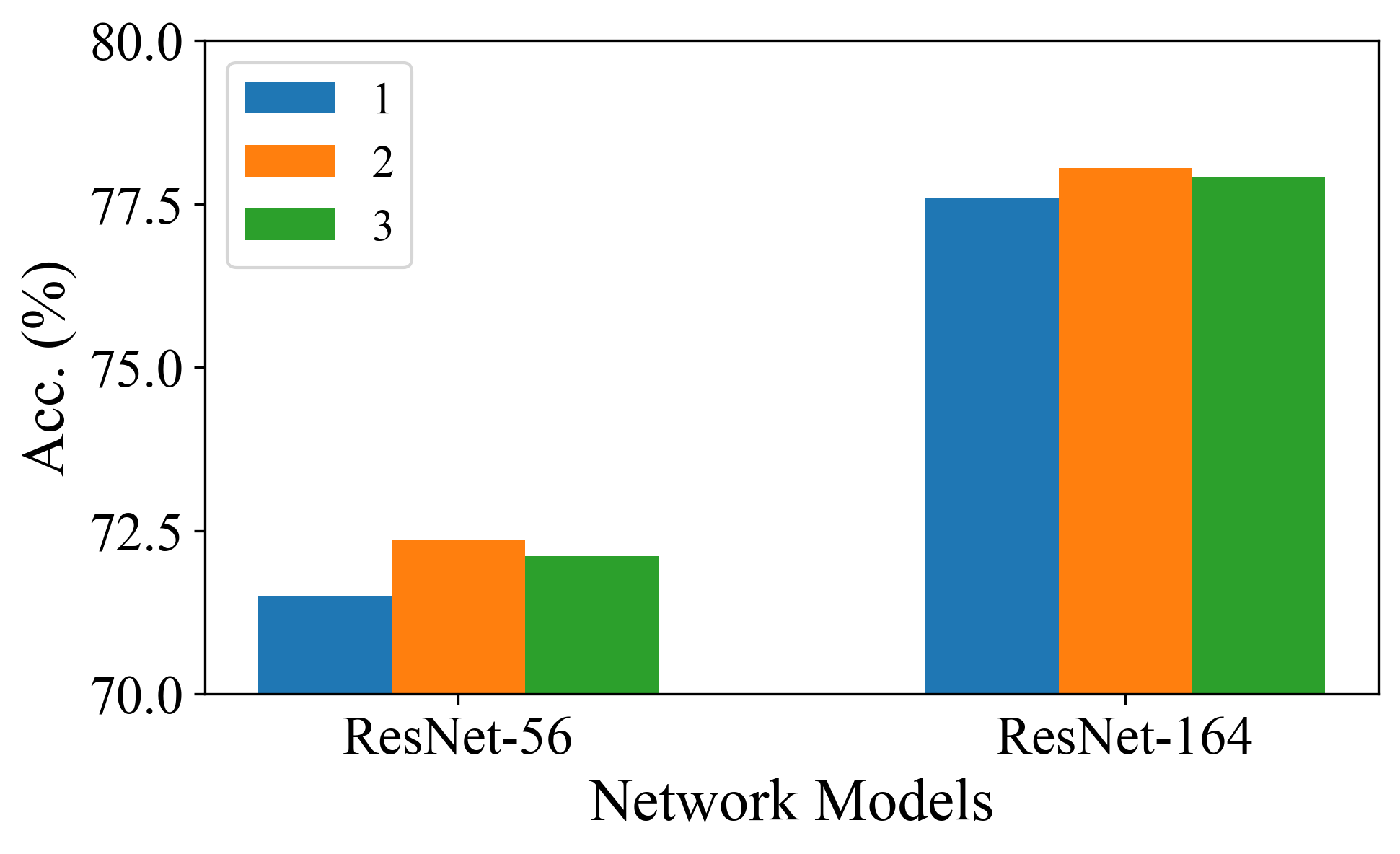}
    \caption{
    Results of multi-hierarchy \method\ with ResNet-56 and ResNet-164 on CIFAR-100. 
    Including a coarse hierarchy for network compression generally works better than the non-hierarchical scheme,
    while increasing the number of hierarchies to 3 gives similar results as the 2-hierarchy scheme.
    }
    \label{fig:multi_hierarchy_CDC}
\end{figure}

We study a 3-hierarchy compression scheme where we learn multiple class hierarchies and insert multiple distillation losses. 
We compare it with the 2-hierarchy scheme from the main paper and the non-hierarchical (only using the fine label) scheme in Fig.~\ref{fig:multi_hierarchy_CDC}.
In the 3-hierarchy scheme, we learn two coarse class labels, $\mathcal{Y}_{c1}$ and $\mathcal{Y}_{c2}$, 
where $\mathcal{Y}_{c2}$ is derived by spectral clustering on the confusion matrix of $\mathcal{Y}_{c1}$, 
and $\mathcal{Y}_{c1}$ is the coarse hierarchy adopted in the 2-hierarchy scheme.
We conduct experiments with ResNet-56/164 on CIFAR-100, where $\mathcal{Y}_{c1}$ has 20 classes and $\mathcal{Y}_{c2}$ has 4 classes. 
We set the watershed layers at the locations of 1/3 and 2/3 of the total layers.
We use $\mathcal{Y}_{c2}$, $\mathcal{Y}_{c1}$, and $\mathcal{Y}_{f}$ to evaluate the channel class discrepancy for the first 1/3, the 1/3 to the 2/3, and the last 1/3 of the layers, 
and we uniformly remove 45\% of channels from each layer.
Two DCA-based distillation losses are inserted at two watershed layers, where the early one use $\mathcal{Y}_{c2}$ to learn weights and the later one use $\mathcal{Y}_{c1}$.
We find both 2-hierarchy and 3-hierarchy schemes achieve better performance than the non-hierarchical one, 
while using 3 hierarchies gives similar results as the 2-hierarchy scheme.

\begin{figure*}[b]
    \centering
    \includegraphics[width=0.9\textwidth]{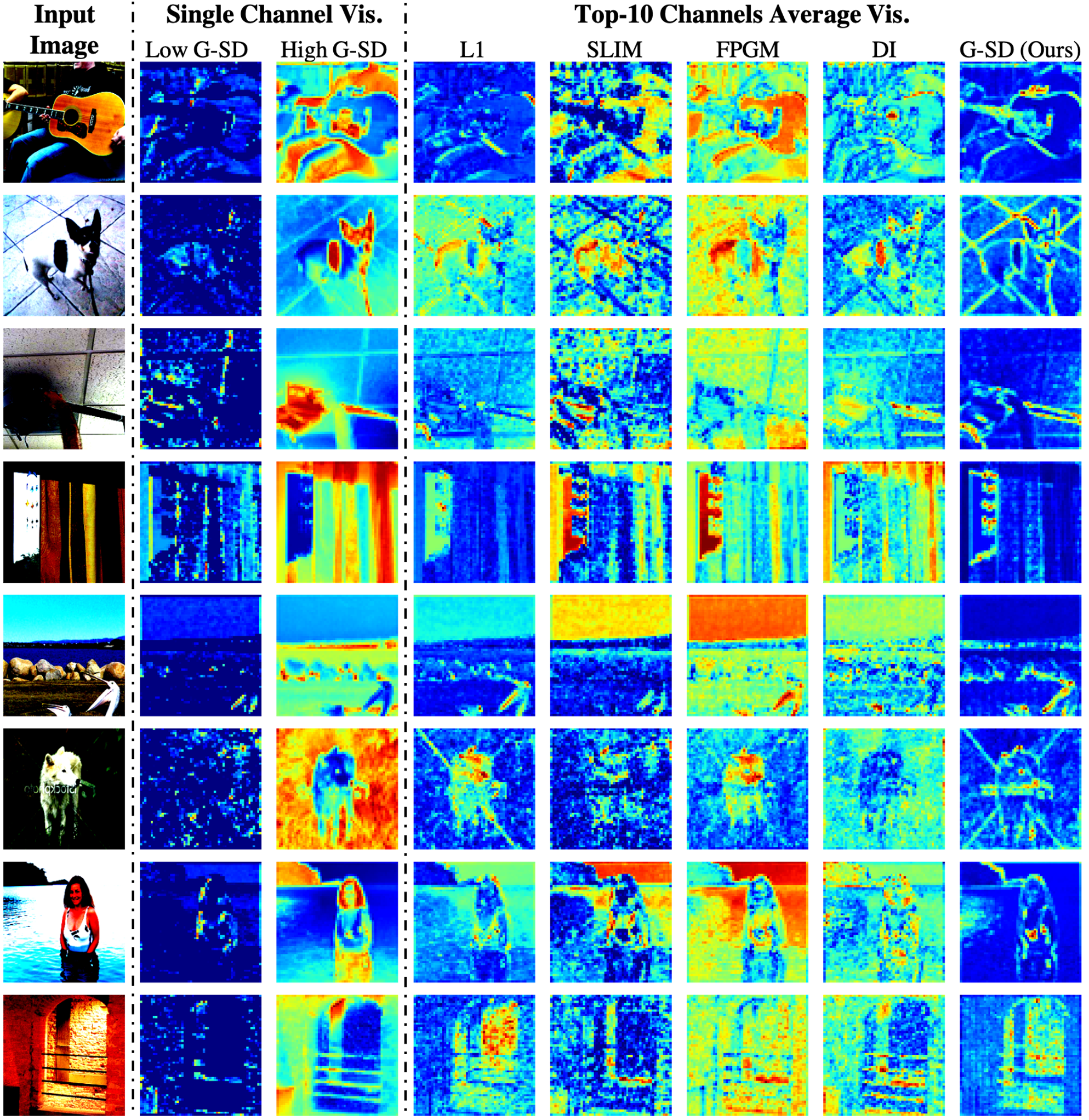}
    
    \caption{
       More visualizations of the channel selection analysis. 
    {\bf Col. 1:} Input images. 
    {\bf Col. 2-3}: Channels with low and high \metric\ values. 
    The low \metric\ channel generates responses that are indistinguishable to different classes, while the high one produces informative activations.
    {\bf Col. 4-8}: Average responses of the top-10 channels selected by different metrics. From left to right, the metrics are: $\ell$1-norm~\cite{li2016pruning}, batch-norm scaling factor~\cite{liu2017learning}, filter's geometric median~\cite{he2019filter}, DI~\cite{kung2019methodical}, and \metric.
    In general, the channels picked by G-SD preserves the most classification information.
    }
    
    \label{fig:Metric_Comparison}
\end{figure*}

\section{More Visualizations}\label{sec:visual}

We repeat the channel selection analysis in Sec.~5 of the main paper with additional images from different classes (not shown in the main paper) in Fig.~\ref{fig:Metric_Comparison}.
Visually speaking, we again observe that our \metric\ selects better class-discriminative channels compared to other methods.

\end{document}